\pdfoutput=1
\documentclass[11pt]{article}

\usepackage{algorithm}
\usepackage{algpseudocode}
\usepackage{booktabs}
\usepackage{multirow}
\usepackage{url}
\usepackage{authblk}
\usepackage{amsmath,amssymb,amsthm}
\usepackage{graphicx}
\usepackage[table]{xcolor}
\usepackage{float} 
\usepackage[numbers,sort&compress]{natbib}
\usepackage[margin=1in]{geometry}

\begin{document}

\title{Delta Score Matters! \\ Spatial Adaptive Multi Guidance in Diffusion Models}
\author[1]{Haosen Li$^{*}$}
\author[1]{Wenshuo Chen$^{*, \ddagger}$}
\author[2]{Lei Wang}
\author[1]{Shaofeng Liang}
\author[1]{Bowen Tian}
\author[1]{Soning Lai}
\author[1]{Yutao Yue$^{\dagger}$}

\affil[1]{The Hong Kong University of Science and Technology (Guangzhou)}
\affil[2]{Griffith University \& Data61/CSIRO}
\affil[]{\small $^{*}$Equal contribution. $^\ddagger$ Project Leader. $^{\dagger}$Corresponding author: yutaoyue@hkust-gz.edu.cn}
\date{}
\maketitle

\begin{figure}[t]
  \includegraphics[width=0.98\textwidth]{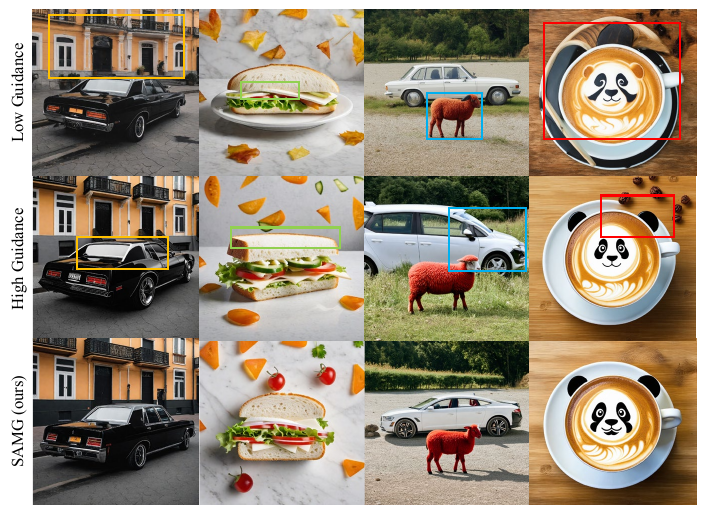}
  \caption{Visualizing the ``detail-artifact dilemma'' in Classifier-Free Guidance and our solution. \textbf{Top (Low Guidance):} Insufficient uniform scales fail to inject adequate semantics, leading to incomplete denoising. \textbf{Middle (High Guidance):} Blindly increasing the global scale enforces semantic alignment but triggers severe manifold deviation, causing color over-saturation, over-exposure, and structural collapse (highlighted in colored boxes). \textbf{Bottom (SAMG, ours):} By dynamically modulating the guidance scale based on spatial energy, our method successfully breaks this dilemma. SAMG marries the structural integrity of low guidance with the rich semantic injection of high guidance, yielding high-fidelity, artifact-free generations.}
  \label{fig:teaser}
\end{figure}

\begin{abstract}
Diffusion models have achieved remarkable success in synthesizing complex static and temporal visuals, a breakthrough largely driven by Classifier-Free Guidance (CFG). However, despite its pivotal role in aligning generated content with textual prompts, standard CFG relies on a globally uniform scalar. This homogeneous amplification traps models in a well-documented "detail-artifact dilemma": low guidance scales fail to inject intricate semantics, while high scales inevitably cause structural degradation, color over-saturation, and temporal inconsistencies in videos. In this paper, we expose the physical root of this flaw through the lens of differential geometry. By analyzing Tweedie's Formula, we reveal that CFG intrinsically performs a tangential linear extrapolation. Because the natural data manifold is highly curved, this uniform linear step introduces a severe orthogonal deviation. To keep the generation trajectory safely bounded, we formulate a theoretical upper bound for spatial and adaptive guidance. Based on these geometric insights, we propose Spatial Adaptive Multi Guidance (SAMG), a training-free and virtually zero-cost sampling algorithm. SAMG dynamically computes point-wise conditional guidance energy, applying a conservative minimum scale to high-energy boundary regions to preserve delicate micro-textures, while deploying an aggressive maximum scale in low-energy regions to maximize semantic injection. Extensive experiments across diverse image (SD 1.5, SDXL, SD3.5 Medium) and video (CogVideoX, ModelScope) architectures demonstrate that SAMG effectively resolves the detail-artifact dilemma, achieving superior semantic alignment, structural integrity, and temporal smoothness without any computational overhead.
\end{abstract}

\section{Introduction}
\label{sec:intro}

Diffusion models have redefined the boundaries of visual generation, encompassing both text-to-image (T2I) and text-to-video (T2V) synthesis ~\cite{ho2020denoisingdiffusionprobabilisticmodels, song2021scorebasedgenerativemodelingstochastic, song2022denoisingdiffusionimplicitmodels, lipman2023flowmatchinggenerativemodeling, liu2022flowstraightfastlearning, rombach2022highresolutionimagesynthesislatent, podell2023sdxlimprovinglatentdiffusion}. They have unlocked extraordinary capabilities in generating high-fidelity, compositionally diverse, and temporally coherent visual content. The cornerstone of this generative breakthrough is Classifier-Free Guidance (CFG) \cite{ho2022classifierfreediffusionguidance}, an elegantly simple yet potent conditioning mechanism. By linearly extrapolating the generative trajectory away from the unconditional prior and towards the text-driven estimate, CFG dramatically amplifies text-alignment. This operation has cemented CFG as the ubiquitous standard across modern generative architectures, seamlessly bridging classic U-Net-based frameworks (e.g., SD 1.5 \cite{rombach2022highresolutionimagesynthesislatent}, SDXL \cite{podell2023sdxlimprovinglatentdiffusion}), recent flow-matched Diffusion Transformers (e.g., SD3.5 \cite{esser2024scalingrectifiedflowtransformers}), and emerging video diffusion models.

\begin{figure}[t]
  \centering 
  \includegraphics[width=\linewidth]{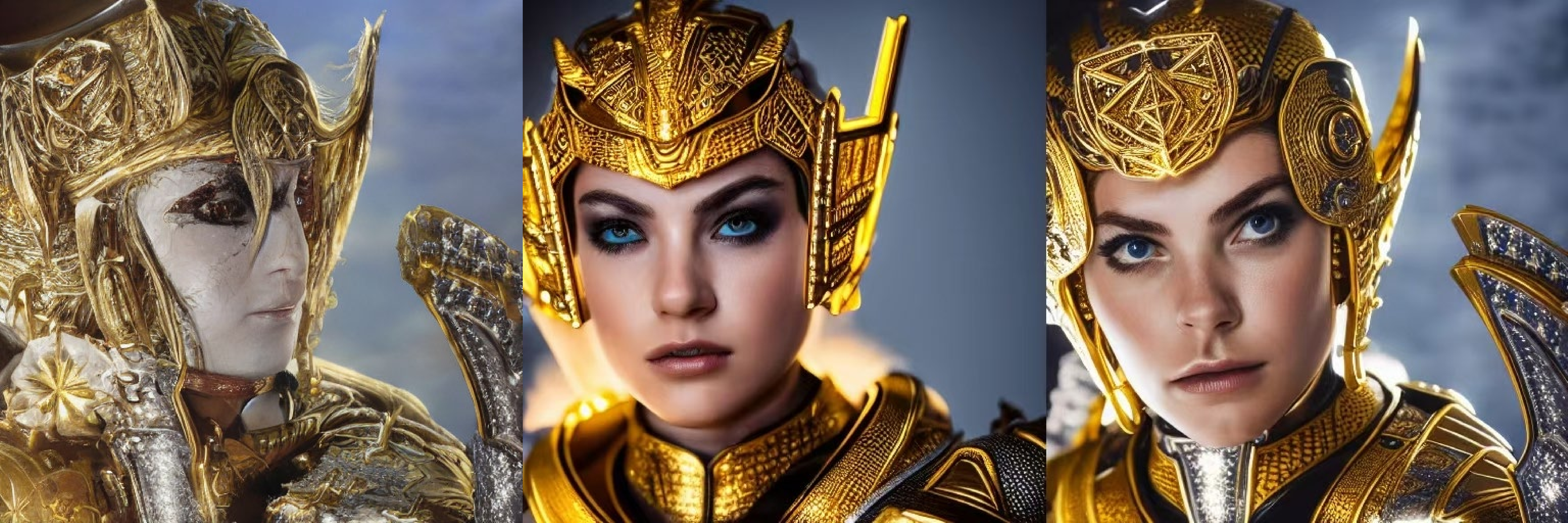}
  \caption{Qualitative comparison of Conditional generation, High CFG, and SAMG. Prompt: ``close up portrait of a futuristic female knight wearing intricate golden armor, magic runes, exquisite skin''. Compared to High CFG, SAMG preserves semantic and textural fidelity while reducing false colors and over-saturation.}
  \label{fig:samg_knight_comparison}
\end{figure}

However, despite its empirical success, the formulation of CFG harbors a fundamental flaw: it applies a \textit{globally uniform scalar} across the entire spatial and temporal resolution of the latent variable. This homogeneous amplification inevitably traps models in a well-documented ``detail-artifact dilemma'' \cite{saharia2022photorealistictexttoimagediffusionmodels, si2023freeufreelunchdiffusion}. As specifically illustrated in Figure \ref{fig:samg_knight_comparison} (Left) and the top row of Figure \ref{fig:teaser}, the standard conditional generation often already successfully establishes the majority of fundamental structural features. To inject missing intricate semantics or force strict text alignment, users are often compelled to drastically increase the global guidance scale. Paradoxically, because this amplification is applied blindly across all regions, it inevitably destroys the local structural integrity that was already well-formed. As shown in Figure \ref{fig:samg_knight_comparison} (Middle) and the middle row of Figure \ref{fig:teaser}, this homogeneous push strictly enforces semantic alignment but simultaneously triggers severe color over-saturation, regional over-exposure, and structural collapse. This stark contrast powerfully motivates the necessity of a spatially varying guidance strategy rather than a rigid global scalar.

To break this bottleneck, we look under the hood of the guidance evolution process through the lens of differential geometry. According to the classical Tweedie's Formula from empirical Bayes ~\cite{Good1992, 6796562}, we reveal that CFG intrinsically performs an explicit \textit{tangential linear extrapolation} on the estimated clean data manifold ~\cite{chung2024cfgmanifoldconstrainedclassifierfree}. Because the true natural visual manifold is highly curved and complex ~\cite{pope2021intrinsicdimensionimagesimpact}, executing a uniform linear step inevitably introduces an orthogonal deviation from the manifold. This theoretical insight exposes the physical root of high-CFG artifacts: pushing homogeneously across all pixels forces high-frequency, high-energy regions (e.g., edges, micro-textures) completely off the data manifold, causing structural and temporal degradation ~\cite{chung2024cfgmanifoldconstrainedclassifierfree, si2023freeufreelunchdiffusion}.

Motivated by this geometric perspective, we formulate a theoretical upper bound for the guidance scale, proving that to keep the generation trajectory safely bounded on the manifold ~\cite{karras2022elucidatingdesignspacediffusionbased, debortoli2022riemannianscorebasedgenerativemodelling}, the ideal guidance scale must not be a global scalar. Instead, it should be a dynamically modulated spatial field that is inversely proportional to the local guidance energy and manifold curvature. Since computing high-dimensional exact curvature during inference is computationally intractable and prone to numerical singularities ~\cite{song2019slicedscorematchingscalable}, we introduce formal mathematical relaxations—specifically, Taylor expansion and affine reparameterization ~\cite{bao2022analyticdpmanalyticestimateoptimal}—to derive a highly efficient approximation.

Based on these theoretical underpinnings, we propose \textbf{Spatial Adaptive Multi Guidance (SAMG)}, a virtually zero-cost, plug-and-play diffusion sampling method. SAMG dynamically calculates the pixel-wise conditional guidance energy $|| \Delta \epsilon ||^2$ at each timestep. It applies a conservative minimum guidance scale ($\omega_{\min}$) to high-energy boundary regions to strictly preserve structural integrity, while deploying an aggressive maximum guidance scale ($\omega_{\max}$) to low-energy regions to maximize semantic injection. As demonstrated in Figure \ref{fig:teaser}, SAMG resolves the detail-artifact dilemma, preserving micro-textures and temporal smoothness while ensuring high semantic alignment.

In summary, the main contributions of this work are three-fold:
\begin{itemize}
    \item \textbf{Theoretical Insight:} We establish a novel geometric perspective on Classifier-Free Guidance, demonstrating that uniform tangential linear extrapolation causes manifold deviation. We mathematically formulate the theoretical upper bound for spatial-adaptive guidance.
    \item \textbf{Methodological Innovation:} We propose SAMG, a training-free, zero-cost sampling algorithm that translates the theoretical manifold constraints into a practical, pixel-wise adaptive modulation mapping applicable to images and videos.
\item \textbf{Empirical Superiority:} Extensive experiments across static image benchmarks (Pick-a-Pic \cite{kirstain2023pickapicopendatasetuser}, DrawBench \cite{saharia2022photorealistictexttoimagediffusionmodels}, MS-COCO 2017 \cite{lin2015microsoftcococommonobjects}, GenEval \cite{ghosh2023genevalobjectfocusedframeworkevaluating}) and video generation benchmarks (ChronoMagic-Bench \cite{yuan2024chronomagic}) using diverse architectures (SD 1.5 \cite{rombach2022highresolutionimagesynthesislatent}, SDXL \cite{podell2023sdxlimprovinglatentdiffusion}, SD3.5 Medium \cite{esser2024scalingrectifiedflowtransformers}, CogVideoX \cite{yang2024cogvideox}, and ModelScope \cite{wang2023modelscope}) demonstrate that SAMG significantly and consistently enhances generation quality, temporal consistency, human preference scores, and text-alignment, without incurring computational overhead.
\end{itemize}

\section{Preliminaries}
\label{sec:preliminaries}

In this section, we briefly review the fundamental concepts that form the basis of our method, specifically unifying the sampling mechanisms of standard latent diffusion models and recent flow-matching architectures, alongside the conditioning techniques used for text-driven visual generation.

\subsection{Latent Diffusion and Flow Models}
\label{subsec:diffusion_flow}

Generative modeling has rapidly evolved from score-based diffusion models to continuous-time flow matching. Both paradigms can be unified under the framework of transforming a simple prior distribution $\mathbf{z}_T \sim \mathcal{N}(\mathbf{0}, \mathbf{I})$ into the complex data distribution $\mathbf{z}_0$ via a deterministic sequence of updates in the latent space.

\textbf{Standard Diffusion Models} (e.g., SD 1.5, SDXL) \cite{rombach2022highresolutionimagesynthesislatent, podell2023sdxlimprovinglatentdiffusion} gradually add Gaussian noise to the data, governed by a strictly decreasing cumulative variance schedule $\bar{\alpha}_t$. A neural network, parameterized as $\epsilon_\theta(\mathbf{z}_t, t)$, is trained to predict the injected noise. During inference, deterministic samplers like Denoising Diffusion Implicit Models (DDIM) \cite{song2022denoisingdiffusionimplicitmodels} solve the reverse probability flow to generate samples. Given a noisy latent $\mathbf{z}_t$, the previous step $\mathbf{z}_{t-1}$ is computed as:
\begin{equation}
\label{eq:ddim_step}
    \mathbf{z}_{t-1} = \sqrt{\bar{\alpha}_{t-1}} \underbrace{\left( \frac{\mathbf{z}_t - \sqrt{1 - \bar{\alpha}_t} \epsilon_\theta(\mathbf{z}_t, t)}{\sqrt{\bar{\alpha}_t}} \right)}_{\text{predicted } \hat{\mathbf{z}}_0} + \underbrace{\sqrt{1 - \bar{\alpha}_{t-1}} \epsilon_\theta(\mathbf{z}_t, t)}_{\text{direction to } \mathbf{z}_t}
\end{equation}

\textbf{Flow Matching Models} (e.g., SD3.5, CogVideoX) \cite{esser2024scalingrectifiedflowtransformers} adopt a Rectified Flow paradigm. Instead of predicting noise, the network estimates a velocity field $\mathbf{v}_\theta(\mathbf{z}_t, t)$ that transports the probability mass along linear trajectories. The sampling is performed via Ordinary Differential Equation (ODE) solvers, such as the Euler method:
\begin{equation}
\label{eq:euler_step}
    \mathbf{z}_{t-\Delta t} = \mathbf{z}_t - \Delta t \cdot \mathbf{v}_\theta(\mathbf{z}_t, t)
\end{equation}

Despite the mathematical differences, both paradigms iteratively construct the data manifold using a predicted directional vector field. For notational simplicity and theoretical continuity, we will predominantly use $\epsilon_\theta$ in our geometric formulations to represent the general neural update direction (denoting either noise or velocity), and abstract the sampling update as a generalized function $\text{Solver\_Step}(\cdot)$.

\subsection{Classifier-Free Guidance}
\label{subsec:cfg}

To control the generation process using user-provided conditions $c$ (such as text prompts), modern architectures predominantly employ Classifier-Free Guidance (CFG) \cite{ho2022classifierfreediffusionguidance} due to its simplicity and superior semantic alignment.

CFG implicitly integrates the conditional and unconditional distributions of a generative model. During training, the conditioning signal $c$ is randomly dropped (replaced with a null token $\emptyset$) with a predefined probability. This allows a single neural network to learn both the conditional estimate $\epsilon_\theta(\mathbf{z}_t, t, c)$ and the unconditional estimate $\epsilon_\theta(\mathbf{z}_t, t, \emptyset)$.

During the sampling phase, CFG modifies the model's prediction by linearly extrapolating in the direction of the conditional prediction. The modified estimate $\tilde{\epsilon}_\theta(\mathbf{z}_t, t, c)$ is defined as:
\begin{equation}
\label{eq:cfg}
    \tilde{\epsilon}_\theta(\mathbf{z}_t, t, c) = \epsilon_\theta(\mathbf{z}_t, t, \emptyset) + \omega \cdot \Big( \epsilon_\theta(\mathbf{z}_t, t, c) - \epsilon_\theta(\mathbf{z}_t, t, \emptyset) \Big)
\end{equation}
where $\omega \ge 1$ is the guidance scale. The difference term acts as an implicit classifier gradient that pushes the sample closer to the data manifold aligned with the condition $c$. Crucially, the predicted noise $\epsilon_\theta(\mathbf{z}_t, t, c)$ is intrinsically equivalent to the scaled conditional score function $-\sqrt{1 - \bar{\alpha}_t} \nabla_{\mathbf{z}_t} \log p_t(\mathbf{z}_t | c)$. This fundamental connection bridges the empirical CFG formulation with the geometric perspective of score matching, which we systematically explore in Section \ref{sec:methodology}.

\section{Methodology}
\label{sec:methodology}

While CFG significantly enhances semantic alignment in text-to-visual generation, its globally uniform scalar formulation overlooks the highly non-linear topological characteristics of the target data manifold. This results in a well-documented quality-diversity trade-off: low guidance scales fail to inject complex semantics, whereas high scales inevitably cause over-saturation and structural artifacts. 

To systematically address this limitation, we re-examine the guidance evolution process through the lens of differential geometry. We formulate a theoretical upper bound for the guidance scale and propose Spatial Adaptive Multi Guidance (SAMG) as a tractable, zero-cost approximation.

\begin{algorithm}[t]
    \caption{Spatial Adaptive Multi Guidance}
    \label{alg:samg}
    \begin{algorithmic}[1]
        \Require Text prompt $c$, Null condition $\emptyset$, Number of timesteps $N$
         Lower guidance bound $\omega_{\min}$, Upper guidance bound $\omega_{\max}$
        \For{$t = N$ \textbf{down to} $1$}
            \State $\epsilon_u = \epsilon_\theta(\mathbf{z}_t, t, \emptyset)$ 
            \State $\epsilon_c = \epsilon_\theta(\mathbf{z}_t, t, c)$
            \State $\Delta \epsilon_t = \epsilon_c - \epsilon_u$
            \State $E_t = \text{Mean}(\Delta \epsilon_t^2, \text{dim}=\text{C})$ 
            \State $\hat{E}_t = \frac{E_t - \min(E_t)}{\max(E_t) - \min(E_t) + \tau}$ 
            \State $\boldsymbol{\Omega}_{map} = \omega_{\max} - \hat{E}_t \cdot (\omega_{\max} - \omega_{\min})$
            \State $\tilde{\epsilon}_{SAMG} = \epsilon_u + \boldsymbol{\Omega}_{map} \odot \Delta \epsilon_t$ 
            \State $\mathbf{z}_{t-1} = \text{Solver\_Step}(\tilde{\epsilon}_{SAMG}, t, \mathbf{z}_t)$ 
        \EndFor
        \Ensure Decoded generated visual $\hat{\mathbf{z}}_0$
    \end{algorithmic}
\end{algorithm}

\subsection{CFG as Tangential Extrapolation}
\label{subsec:theoretical_insight}

For notational brevity, let $\epsilon_u \triangleq \epsilon_\theta(\mathbf{z}_t, t, \emptyset)$ and $\epsilon_c \triangleq \epsilon_\theta(\mathbf{z}_t, t, c)$. At an arbitrary denoising timestep $t$, the standard CFG modifies the predicted vector field as follows:
\begin{equation}
    \tilde{\epsilon}_\omega(\mathbf{z}_t) = \epsilon_{u} + \omega \underbrace{\big( \epsilon_{c} - \epsilon_{u} \big)}_{\triangleq \Delta \epsilon_t} = \epsilon_{u} + \omega \Delta \epsilon_t
    \label{eq:cfg_base}
\end{equation}
where $\Delta \epsilon_t$ represents the conditional guidance direction, \textbf{which we formally define as the Delta Score} in this work.

To unveil the geometric significance behind this operation, we trace back to the foundational statistical principles. According to the classical \textbf{Tweedie's Formula} \cite{Good1992, 6796562} from empirical Bayes, the posterior mean of the clean data $\mathbf{z}_0$, given a Gaussian-corrupted observation $\mathbf{z}_t$, is strictly determined by the true marginal score function $\nabla_{\mathbf{z}_t} \log p_t(\mathbf{z}_t)$:
\begin{equation}
    \mathbb{E}[\mathbf{z}_0 | \mathbf{z}_t] = \frac{1}{\sqrt{\bar{\alpha}_t}} \Big( \mathbf{z}_t + (1-\bar{\alpha}_t) \nabla_{\mathbf{z}_t} \log p_t(\mathbf{z}_t) \Big)
    \label{eq:original_tweedie}
\end{equation}

In the context of score-based generative models, substituting the score field with our CFG-modified neural prediction $\tilde{\epsilon}_\omega \approx -\sqrt{1-\bar{\alpha}_t} \nabla_{\mathbf{z}_t} \log p_t(\mathbf{z}_t)$ projects the current latent state onto the estimated conditional data manifold $\mathcal{M}$:
\begin{equation}
    \hat{\mathbf{z}}_0(\omega) = \frac{\mathbf{z}_t - \sqrt{1-\bar{\alpha}_t} \tilde{\epsilon}_\omega}{\sqrt{\bar{\alpha}_t}} = \underbrace{\frac{\mathbf{z}_t - \sqrt{1-\bar{\alpha}_t} \epsilon_u}{\sqrt{\bar{\alpha}_t}}}_{\triangleq \hat{\mathbf{z}}_{0, u}} - \omega \frac{\sqrt{1-\bar{\alpha}_t}}{\sqrt{\bar{\alpha}_t}} \Delta \epsilon_t
    \label{eq:tweedie_projection}
\end{equation}

\textit{Remark:} For flow-based models, an analogous projection occurs during the Euler integration $\mathbf{z}_{t-\Delta t} = \mathbf{z}_t - \Delta t (\mathbf{v}_u + \omega \Delta \mathbf{v}_t)$, which equally acts as a linear step scaled by $\omega \Delta \mathbf{v}_t$.

Both formulations reveal a core physical mechanism: \textbf{CFG intrinsically performs an explicit tangential linear extrapolation}, starting from the unconditional anchor ($\hat{\mathbf{z}}_{0, u}$ or $\mathbf{z}_{t, u}$) and moving aggressively along the tangent space towards the direction of $-\Delta \epsilon_t$. 

To quantify the magnitude of this local guidance, we define the spatial \textit{guidance energy} $E_t(x)$ at spatial coordinate $x$ as the channel-wise mean of the squared difference:
\begin{equation}
    E_t(x) = \frac{1}{C} \|\Delta \epsilon_t(x)\|_2^2
\end{equation}
where $C$ is the channel dimension. This energy term intrinsically governs the effective physical extrapolation distance in the feature space during sampling.

\subsection{Theoretical Bound: Manifold Deviation}
\label{subsec:manifold_deviation}

Real-world natural visuals reside on a highly non-linear Riemannian manifold $\mathcal{M} \subset \mathbb{R}^D$ embedded within the high-dimensional latent space. Executing a linear extrapolation step along the tangent space $T_x \mathcal{M}$ inherently introduces an orthogonal deviation from the true manifold.

\noindent\textbf{A Numerical Illustration.} To concretely grasp this geometric constraint, consider the guidance vectors at two distinct spatial locations: $v_1 = (0.1, 0.2, -0.1)$ in a low-frequency smooth region, and $v_2 = (1.5, 3.0, -1.5)$ at a high-frequency texture boundary. It is worth noting that the roles of the guidance direction and the local guidance magnitude are not redundant. Although both vectors share identical optimization directions towards the text condition, their magnitudes differ drastically. Applying a fixed high scale (e.g., $\omega = 7.0$) yields a moderate, safe tangential step of $(0.7, 1.4, -0.7)$ for $v_1$. However, the exact same global scalar forces a massive extrapolation step of $(10.5, 21.0, -10.5)$ for $v_2$, which inevitably overshoots the highly curved manifold boundaries. From a geometric perspective, dynamically bounding the local magnitude facilitates the model to more stably preserve crucial micro-textures without causing structural degradation.

\begin{figure}[t]
 \centering
 \includegraphics[width=\textwidth]{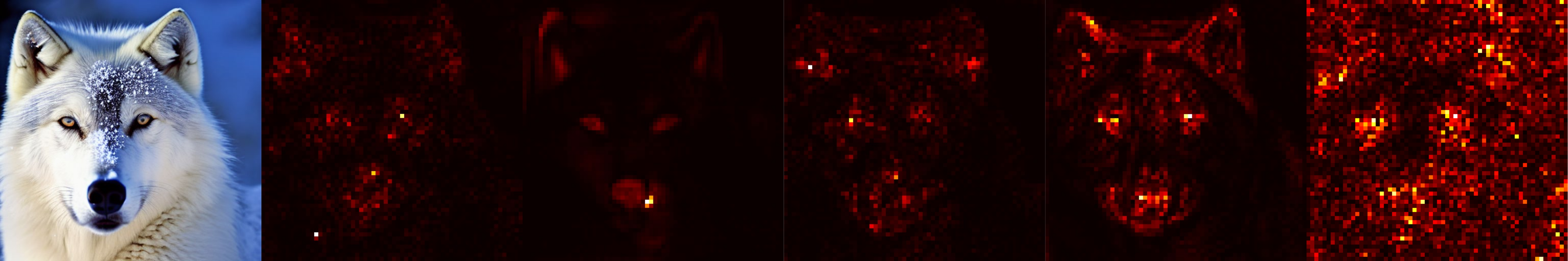}
    \caption{Visualization of the generated image and the evolution of spatial guidance energy maps $E_t(x)$ during the denoising process. From left to right, the first panel shows the generated image, while the remaining panels depict energy maps at different timesteps. Bright regions correspond to high-frequency and structurally complex areas, such as the wolf's eyes, nose contour, and fine snow particles on its fur. These visualizations confirm that SAMG accurately identifies delicate micro-structures and applies adaptive guidance modulation to preserve local details.}
 \label{fig:energy_map}
\end{figure}

Mathematically, at a given spatial coordinate $x$, let its unconditional anchor be the starting point $\mathbf{p} \in \mathbb{R}^C$ on the local data sub-manifold. Let $\mathbf{v} = -\Delta \epsilon_t(x) / \|\Delta \epsilon_t(x)\|_2$ denote the normalized extrapolation direction. The ideal generation trajectory bounded safely on the manifold is described by the true geodesic curve $\gamma(s) = \exp_{\mathbf{p}}(s \mathbf{v})$, where the effective physical step size $s$ is governed by $s = \omega \frac{\sqrt{1-\bar{\alpha}_t}}{\sqrt{\bar{\alpha}_t}} \|\Delta \epsilon_t(x)\|_2 = \omega \frac{\sqrt{1-\bar{\alpha}_t}}{\sqrt{\bar{\alpha}_t}} \sqrt{C \cdot E_t(x)}$. Thus, $s$ is directly proportional to the scaled guidance energy: $s \propto \omega \sqrt{E_t(x)}$. Standard CFG, however, strictly reaches the linearly extrapolated coordinate $y = \mathbf{p} + s \mathbf{v}$. The discrepancy between $y$ and $\mathcal{M}$ constitutes structural degradation.

\noindent\textbf{Proposition 1 (Manifold Deviation Bound).} \textit{For a sufficiently small step size $s$, let $\kappa(x)$ be the local extrinsic normal curvature along direction $\mathbf{v}$, governed by the manifold's Second Fundamental Form. The deviation error $\mathcal{E}_{dev}(x) = \| y - \gamma(s) \|_2$ is bounded by:
\begin{equation}
    \mathcal{E}_{dev}(x) \approx \frac{1}{2} \kappa(x) s^2 \propto \frac{1}{2} \kappa(x) \omega^2 E_t(x)
    \label{eq:deviation_error}
\end{equation}
To strictly bound the generation trajectory within a safe $\delta$-neighborhood of the natural visual manifold ($\mathcal{E}_{dev}(x) \le \delta$), the optimal pixel-wise adaptive guidance scale $\omega_{ideal}(x)$ must satisfy:}
\begin{equation}
    \omega_{ideal}(x) \le \sqrt{\frac{2\delta}{\kappa(x) \cdot c_t E_t(x)}} \propto \frac{1}{\sqrt{\kappa(x) E_t(x)}}
    \label{eq:omega_ideal}
\end{equation}
\textit{where $c_t$ is a timestep-dependent deterministic proportionality constant (e.g., $c_t = C \frac{1-\bar{\alpha}_t}{\bar{\alpha}_t}$ for Tweedie projection in diffusion models, or $c_t = C (\Delta t)^2$ for Euler integration in flow models).}

Proposition 1 formally validates our physical observation: the ideal guidance scale is not a global scalar, but a dynamically modulated spatial field inversely proportional to the square root of the local guidance energy and curvature.

\subsection{SAMG: Tractable Approximation}
\label{subsec:samg_approximation}

Computing the exact high-dimensional curvature $\kappa(x)$ during inference is computationally intractable, and the theoretical inverse square root formulation exhibits severe numerical singularities as $E_t(x) \to 0$. To derive a practical and efficient algorithm, we propose Spatial Adaptive Multi Guidance (SAMG) based on two formal relaxations.

\noindent\textbf{Assumption 1 (Bounded Curvature).}
\textit{We assume the extrinsic curvature of the natural data manifold $\mathcal{M}$ is upper-bounded globally. Under this condition, the manifold deviation risk is dominantly governed by the local guidance energy $E_t(x)$.}

Based on Assumption 1, the optimal scale simplifies to $\omega \propto 1/\sqrt{E_t(x)}$. To bypass the singularity, we perform a first-order Taylor expansion at a non-zero local operating point $\eta_0$:
\begin{equation}
    \frac{1}{\sqrt{E_t(x)}} \approx \frac{1}{\sqrt{\eta_0}} - \frac{1}{2 \eta_0^{3/2}} \big(E_t(x) - \eta_0\big) = C_1 - C_2 \cdot E_t(x)
    \label{eq:taylor_expansion}
\end{equation}
where $C_1$ and $C_2$ are positive constants. Crucially, since the target function $f(E) = E^{-1/2}$ is strictly convex for $E > 0$ (its second derivative is strictly positive), its first-order Taylor expansion inherently serves as a strict global lower bound: $C_1 - C_2 \cdot E_t(x) \le E_t(x)^{-1/2}$. This geometric property guarantees that our negative-slope linear relaxation is not merely a heuristic approximation, but a mathematically conservative boundary that strictly respects the theoretical manifold upper bound without risking over-extrapolation.

Since the exact constants $C_1$ and $C_2$ are analytically intractable in the high-dimensional latent space, we relax this theoretical negative-slope linear boundary into a practical empirical mapping. Specifically, we introduce an \textbf{Affine Reparameterization} strategy operating within a predefined safety interval $[\omega_{\min}, \omega_{\max}]$. We first normalize the dynamic spatial energy at step $t$ to $\hat{E}_t(x) \in [0,1]$:
\begin{equation}
    \hat{E}_t(x) = \frac{E_t(x) - \min(E_t)}{\max(E_t) - \min(E_t) + \tau}
\end{equation}
where $\tau=10^{-8}$ is for numerical stability, and the $\min/\max$ operations are performed spatially across the current timestep.

To robustly instantiate the negative-slope linear bounding derived in Equation (\ref{eq:taylor_expansion}), we map the normalized energy to construct the adaptive spatial guidance field $\boldsymbol{\Omega}_{map}(x)$:
\begin{equation}
    \boldsymbol{\Omega}_{map}(x) = \omega_{\max} - \hat{E}_t(x) \cdot (\omega_{\max} - \omega_{\min})
    \label{eq:w_map}
\end{equation}

This reparameterization assigns a conservative minimum scale ($\omega_{\min}$) to high-energy boundary regions ($\hat{E}_t \to 1$) to strictly preserve structural integrity, while deploying an aggressive maximum scale ($\omega_{\max}$) to low-energy regions ($\hat{E}_t \to 0$) to maximize semantic injection. The final SAMG-modulated score prediction is computed point-wise as $\tilde{\epsilon}_{SAMG} = \epsilon_u + \boldsymbol{\Omega}_{map}(x) \odot \Delta \epsilon_t$. To ensure stringent geometric constraints are not diluted from high-energy boundaries to safer regions, we enforce pixel-wise independent calculations for $E_t(x)$ without any spatial smoothing. The complete zero-cost pixel-wise sampling procedure is summarized in Algorithm \ref{alg:samg}.

\section{Empirical Analysis}
\subsection{Experimental Settings}
\label{subsec:exp_setting}

To evaluate the effectiveness of different guidance strategies, we conduct extensive experiments across multiple models, datasets, and evaluation metrics. The settings are outlined below.

\paragraph{Datasets.}
We evaluate the text-to-image generation performance on four widely adopted benchmarks. 
1) \textbf{Pick-a-Pic}~\cite{kirstain2023pickapicopendatasetuser}: A large-scale dataset of text-to-image prompts collected from real users, which is highly effective for evaluating human preference and diverse generation scenarios. 
2) \textbf{DrawBench}~\cite{saharia2022photorealistictexttoimagediffusionmodels}: A challenging prompt suite designed to test the fundamental capabilities of text-to-image models, including compositionality, spatial relations, and complex text comprehension. 
3) \textbf{GenEval}~\cite{ghosh2023genevalobjectfocusedframeworkevaluating}: An objective, rule-based evaluation benchmark specifically designed to systematically assess fine-grained compositional capabilities, such as object counting, attribute binding, and precise spatial relations. 
4) \textbf{MS-COCO 2017}~\cite{lin2015microsoftcococommonobjects}: A standard benchmark extensively used for zero-shot text-to-image evaluation. We utilize its validation set prompts to measure overall image generation quality and general text-image alignment under standard descriptive conditions. 
Additionally, to evaluate text-to-video generation, we utilize \textbf{ChronoMagic-Bench-150} \cite{yuan2024chronomagic}, a comprehensive benchmark designed to assess temporal consistency, motion naturalness, and prompt adherence in video synthesis.

\paragraph{Evaluation Metrics.}
To provide a holistic assessment of generation quality, human preference, and text-image alignment, we employ a comprehensive suite of automated metrics. 
\textbf{HPSv2}~\cite{wu2023human} (Human Preference Score v2) and \textbf{ImageReward}~\cite{xu2023imagerewardlearningevaluatinghuman} are utilized to measure how well the generated images align with human aesthetic preferences and intent. 
\textbf{Aesthetic Score}~\cite{schuhmann2022laion5bopenlargescaledataset} is used to evaluate the overall visual appeal and artistic quality of the outputs. 
To quantify the foundational image fidelity and diversity relative to the real data distribution, we compute the \textbf{FID}~\cite{Seitzer2020FID} (Fréchet Inception Distance).
\textbf{CLIP Score}~\cite{radford2021learningtransferablevisualmodels} strictly quantifies the semantic consistency between the generated images and the conditioning text prompts. 
Finally, we report \textbf{Top-$\boldsymbol{k}$ Accuracy}~\cite{NIPS2012_c399862d} (specifically Top-1, Top-5, and Top-10) to evaluate the robust success rate of the generated images being ranked as the most preferred and highly aligned outputs within candidate sets. For video generation tasks, we additionally evaluate temporal motion consistency using \textbf{CHScore Flow \cite{yuan2024chronomagic}}, frame-level structural preservation via \textbf{Frame LPIPS \cite{zhang2018perceptual}} and \textbf{Frame SSIM \cite{nilsson2020understandingssim}}, and spatial-temporal semantic alignment using \textbf{CLIP SIM \cite{radford2021learningtransferablevisualmodels}} and \textbf{MTScore CLIP}.

\paragraph{Diffusion Models \& Parameters.}
To demonstrate the generalization of our guidance method, we evaluate across diverse architectures. For \textbf{Stable Diffusion 1.5 (SD 1.5)}~\cite{rombach2022highresolutionimagesynthesislatent} and \textbf{Stable Diffusion XL (SDXL)}~\cite{podell2023sdxlimprovinglatentdiffusion}, we use 50 inference steps with a default global CFG scale~\cite{ho2022classifierfreediffusionguidance} of $7.5$, setting SAMG bounds $[\omega_{\min}, \omega_{\max}]$ to $[5.0, 12.0]$. Notably, when evaluating FID on the \textbf{COCO} dataset, we lower the CFG scale to $2.0$ and set the SAMG bounds to $[1.0, 3.0]$ with a pixel-wise independent mapping (Kernel = 1). For \textbf{Stable Diffusion 3.5 Medium (SD3.5-M)}~\cite{esser2024scalingrectifiedflowtransformers}, we use 28 steps, a default CFG of $4.5$, and SAMG bounds of $[3.0, 7.5]$. For text-to-video generation, we evaluate \textbf{CogVideoX-2B~\cite{yang2024cogvideox}} and \textbf{ModelScope-1.7B~\cite{wang2023modelscope}}. Both use 50 steps, a CFG of $7.5$, and SAMG bounds of $[5.0, 12.0]$, also utilizing Kernel = 1 to prevent energy leakage.

\paragraph{Guidance Methods.}
We compare the standard \textbf{CFG}~\cite{ho2022classifierfreediffusionguidance} baseline against \textbf{Perturbed Attention Guidance (PAG)}~\cite{ahn2024self}, \textbf{CFG++}~\cite{chung2024cfgmanifoldconstrainedclassifierfree}, and \textbf{CFG-Zero}~\cite{fan2025cfgzerostar}. For PAG, we maintain the default CFG scale and empirically set its applied PAG scale to 4.5. Notably, the formulation of CFG++ operates on a distinctly different scale interval; therefore, its baseline global scale is evaluated at 0.6, and when augmented with our SAMG, its adaptive bounds are set to [0.4, 1.0].

Table \ref{tab:combined_results_wide} summarizes the performance of SAMG against baseline guidance strategies on the Pick-a-Pic and DrawBench datasets. A consistent trend across all architectures (SD 1.5, SDXL, and SD3.5-M) is that even when employing the widely accepted optimal global scales (e.g., CFG 7.5 for SD 1.5/SDXL and 4.5 for SD3.5-M), standard CFG encounters a rigid performance ceiling. At these carefully tuned baseline settings, the model struggles to further enhance semantic alignment (CLIP) without compromising structural fidelity and human preference (ImageReward). SAMG effectively breaks this bottleneck. For instance, on SDXL (DrawBench), integrating SAMG with standard CFG elevates the CLIP score to 68.33 and Top-1 accuracy to 82.63\% without sacrificing aesthetic quality. Furthermore, SAMG proves highly complementary to recent advanced methods. When combined with CFG++, SAMG actively recovers the human preference degradation typically caused by CFG++'s rigid formulation (e.g., boosting the SD 1.5 ImageReward from -0.229 to -0.089), achieving near-perfect Top-5 and Top-10 accuracies while maintaining semantic rigor. The consistent gains on the recent SD3.5-M further confirm that our geometric manifold theory is architecturally agnostic.

\begin{table*}[t]
\centering
\caption{Quantitative comparison of different guidance strategies on the \textbf{Pick-a-Pic} and \textbf{DrawBench} datasets. Abbreviations: HPS = HPSv2, IR = ImageReward, Aes. = Aesthetic, T1/T5/T10 = Top-1/5/10 Accuracy. All metrics are higher-better ($\uparrow$).}
\label{tab:combined_results_wide}
\resizebox{\textwidth}{!}{
\begin{tabular}{ll ccccccc @{\hspace{2em}} ccccccc}
\toprule
\multirow{2}{*}{\textbf{Model}} & \multirow{2}{*}{\textbf{Method}} & \multicolumn{7}{c@{\hspace{2em}}}{\textbf{Pick-a-Pic}} & \multicolumn{7}{c}{\textbf{DrawBench}} \\
\cmidrule(lr){3-9} \cmidrule(l){10-16}
& & HPS & IR & Aes. & CLIP & T1 & T5 & T10 & HPS & IR & Aes. & CLIP & T1 & T5 & T10 \\
\midrule

\multirow{5}{*}{\textbf{SD 1.5}} 
& CFG                 & 26.87 & \textbf{0.053} & 5.633 & 66.07 & 88 & 98 & 99 
                      & 28.13 & 0.020 & 5.407 & 64.96 & 75 & 95 & 99 \\
& PAG                 & 26.48 & 0.008 & 5.504 & 66.44 & 86 & 96 & 96 
                      & 27.95 & \textbf{0.022} & 5.264 & \textbf{66.73} & 78 & 94 & 97 \\
& CFG + SAMG          & \textbf{26.98} & 0.043 & \textbf{5.642} & \textbf{66.74} & \textbf{91} & \textbf{100} & \textbf{100} 
                      & \textbf{28.31} & 0.000 & \textbf{5.465} & 65.34 & \textbf{83} & \textbf{98} & \textbf{100} \\
\cmidrule{2-16}
& CFG++               & 26.46 & -0.229 & 5.692 & 66.02 & \textbf{93} & 99 & \textbf{100} 
                      & 27.58 & -0.215 & 5.400 & 64.51 & 77 & \textbf{96} & 98 \\
& CFG++ + SAMG        & \textbf{26.66} & \textbf{-0.089} & \textbf{5.742} & \textbf{66.12} & 90 & \textbf{99} & \textbf{100} 
                      & \textbf{27.87} & \textbf{-0.080} & \textbf{5.477} & \textbf{65.54} & \textbf{79} & 95 & \textbf{99} \\
\midrule

\multirow{5}{*}{\textbf{SDXL}}   
& CFG                 & 28.32 & 0.697 & 6.059 & 70.04 & 87 & 99 & 100 
                      & 28.67 & 0.584 & 5.665 & 67.76 & 82 & 94 & 97 \\
& PAG                 & 28.36 & 0.670 & 6.045 & 69.21 & \textbf{89} & 98 & 98 
                      & 28.82 & 0.587 & 5.660 & 67.83 & 81 & \textbf{95} & 96 \\
& CFG + SAMG          & \textbf{28.70} & \textbf{0.741} & \textbf{6.064} & \textbf{70.46} & 88 & \textbf{100} & \textbf{100} 
                      & \textbf{28.87} & \textbf{0.599} & \textbf{5.666} & \textbf{68.33} & \textbf{83} & 94 & \textbf{97} \\
\cmidrule{2-16}
& CFG++               & 28.39 & 0.756 & 6.147 & 71.51 & 90 & 99 & \textbf{100} 
                      & 29.17 & 0.666 & 5.745 & 69.54 & 78 & 93 & 96 \\
& CFG++ + SAMG        & \textbf{28.56} & \textbf{0.810} & \textbf{6.159} & \textbf{72.08} & \textbf{91} & \textbf{99} & 99 
                      & \textbf{29.44} & \textbf{0.754} & \textbf{5.746} & \textbf{69.63} & \textbf{79} & \textbf{94} & \textbf{96} \\
\midrule

\multirow{4}{*}{\textbf{SD3.5-M}} 
& CFG                 & 28.47 & 0.834 & 5.947 & 68.24 & 87 & 97 & 98 
                      & 29.11 & 0.823 & 5.455 & 70.69 & 78 & \textbf{94} & \textbf{96} \\
& CFG + SAMG          & \textbf{28.66} & \textbf{0.924} & \textbf{5.959} & \textbf{68.74} & \textbf{88} & \textbf{97} & \textbf{98} 
                      & \textbf{29.53} & \textbf{0.968} & \textbf{5.474} & \textbf{71.61} & \textbf{82} & 94 & \textbf{96} \\
\cmidrule{2-16}
& CFG-Zero            & 28.32 & 0.813 & 5.949 & 67.85 & 88 & 97 & 98 
                      & 29.04 & 0.819 & 5.501 & 70.32 & 77 & 93 & \textbf{96} \\
& CFG-Zero + SAMG     & \textbf{28.56} & \textbf{0.849} & \textbf{5.954} & \textbf{68.84} & \textbf{89} & \textbf{98} & \textbf{98} 
                      & \textbf{29.40} & \textbf{0.929} & \textbf{5.546} & \textbf{70.99} & \textbf{80} & \textbf{94} & \textbf{96} \\
\bottomrule
\end{tabular}
}
\end{table*}

\begin{table*}[t]
\centering
\caption{Quantitative comparison of CFG and SAMG on the \textbf{GenEval} and \textbf{COCO2017} benchmarks. For GenEval and CLIPScore, higher is better ($\uparrow$); for FID, lower is better ($\downarrow$). Abbreviations: Sgl = Single Object, Two = Two Objects, Cnt = Counting, Col = Colors, Pos = Position, Attr = Attribute Binding. Best results are highlighted in \textbf{bold}.}
\label{tab:geneval}
\label{tab:coco_eval}
\resizebox{\textwidth}{!}{
\begin{tabular}{@{}ll ccccccc cc@{}}
\toprule
\multirow{2}{*}{\textbf{Model}} 
& \multirow{2}{*}{\textbf{Method}} 
& \multicolumn{7}{c}{\textbf{GenEval} ($\uparrow$)} 
& \multicolumn{2}{c}{\textbf{COCO2017}} \\
\cmidrule(lr){3-9} \cmidrule(l){10-11}
& & All & Sgl & Two & Cnt & Col & Pos & Attr 
& FID $\downarrow$ & CLIPScore $\uparrow$ \\
\midrule

\multirow{2}{*}{\textbf{SD 1.5}} 
& CFG 
& 0.43 & 0.97 & 0.38 & 0.35 & 0.76 & 0.04 & 0.06 
& 19.4704 & 20.21 \\
& \textbf{SAMG} 
& \textbf{0.45} & \textbf{0.98} & \textbf{0.40} & \textbf{0.37} & \textbf{0.78} & \textbf{0.05} & \textbf{0.08} 
& \textbf{19.2591} & \textbf{20.55} \\

\midrule

\multirow{2}{*}{\textbf{SDXL}} 
& CFG 
& 0.55 & 0.98 & 0.74 & 0.39 & 0.85 & 0.15 & 0.23 
& 27.9759 & 19.51 \\
& \textbf{SAMG} 
& \textbf{0.58} & \textbf{0.99} & \textbf{0.76} & \textbf{0.42} & \textbf{0.87} & \textbf{0.18} & \textbf{0.26} 
& \textbf{25.4147} & \textbf{20.14} \\

\midrule

\multirow{2}{*}{\textbf{SD3.5-M}} 
& CFG 
& 0.63 & 0.98 & 0.78 & 0.50 & 0.81 & 0.24 & 0.52 
& 21.9324 & 18.86 \\
& \textbf{SAMG} 
& \textbf{0.66} & \textbf{0.99} & \textbf{0.81} & \textbf{0.53} & \textbf{0.83} & \textbf{0.27} & \textbf{0.56} 
& \textbf{20.8680} & \textbf{19.41} \\

\bottomrule
\end{tabular}
}
\end{table*}

\begin{figure*}[t]
 \centering
 \includegraphics[width=\textwidth]{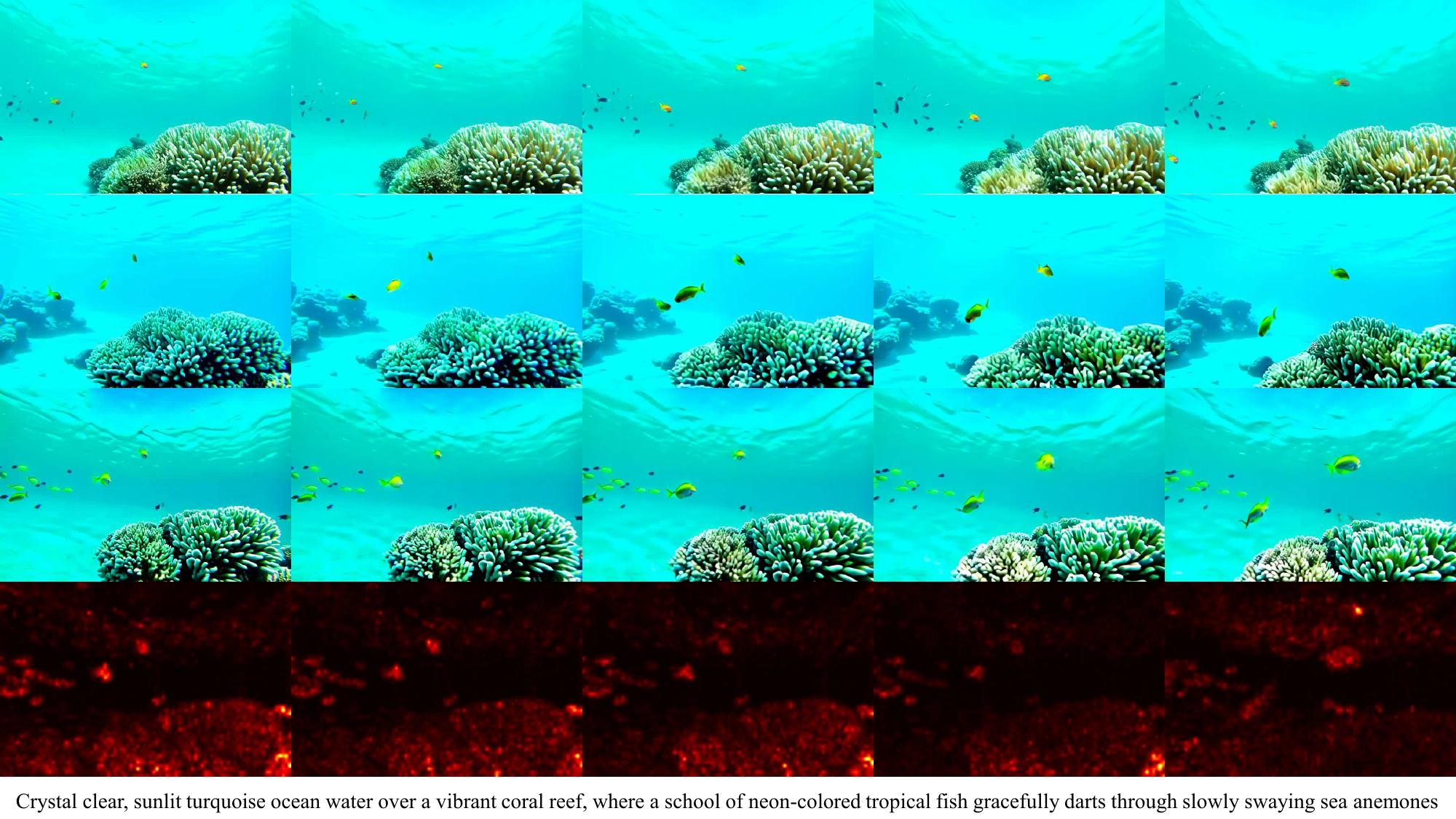}
 \caption{Qualitative comparison of video generation. \textbf{Row 1 (Low CFG):} Preserves background details (ocean ripples) but lacks semantics, yielding blurry fish. \textbf{Row 2 (High CFG):} Enhances semantics but causes temporal discontinuity (abrupt fish count changes) and structural degradation. \textbf{Row 3 (SAMG, ours):} Overcomes this dilemma, synthesizing clear, temporally consistent subjects while preserving fine background details. \textbf{Row 4 (Energy Map):} Visualizes spatial-temporal energy $E_t(x)$, confirming SAMG accurately targets high-frequency local regions in videos.}
 \label{fig:video_comparison}
\end{figure*}

\begin{table*}[t]
\centering
\caption{Quantitative evaluation on ChronoMagic-Bench-150 (CogVideoX-2B and ModelScope-1.7B).}
\label{tab:chrono_eval}
\resizebox{\textwidth}{!}{
\begin{tabular}{llccccc}
\toprule
\textbf{Model} & \textbf{Method} & \textbf{CHScore Flow}$\uparrow$ & \textbf{CLIP SIM}$\uparrow$ & \textbf{Frame LPIPS}$\downarrow$ & \textbf{Frame SSIM}$\uparrow$ & \textbf{MTScore CLIP}$\uparrow$ \\
\midrule
\multirow{2}{*}{CogVideoX-2B} 
& CFG         & 59.93          & 29.00          & 9.68          & 87.07          & 5.38 \\
& SAMG (ours) & \textbf{60.15} & \textbf{29.18} & \textbf{9.35} & \textbf{87.45} & \textbf{5.43} \\
\midrule
\multirow{2}{*}{ModelScope-1.7B} 
& CFG         & 70.47          & 27.79          & 8.12          & 79.13          & 10.56 \\
& SAMG (ours) & \textbf{72.10} & \textbf{27.95} & \textbf{7.35} & \textbf{80.50} & \textbf{10.78} \\
\bottomrule
\end{tabular}
}
\end{table*}

\subsection{Text-to-Image Generation}
\label{subsec:main_experiments}

To further evaluate zero-shot generation quality and standard text-image alignment, we report the Fréchet Inception Distance (FID) and CLIPScore on the MS-COCO 2017 validation set (Table \ref{tab:coco_eval}). The results demonstrate that SAMG consistently outperforms the 
CFG baseline across all tested architectures. Standard guidance typically forces a trade-off where higher semantic alignment degrades image quality. However, SAMG breaks this barrier, achieving lower FID scores (e.g., a significant reduction from 27.9759 to 25.4147 on SDXL) while simultaneously boosting the CLIPScore. 

This indicates that our adaptive bounding strictly preserves the natural image distribution while enhancing text adherence.

Moreover, we assess the fine-grained compositional generation capabilities of our method using the GenEval benchmark (Table \ref{tab:geneval}). The integration of SAMG yields comprehensive improvements across all evaluated models and granular metrics. Beyond single-object generation, SAMG exhibits particular strength in complex compositional scenarios that require precise spatial understanding and attribute isolation. For instance, on SD3.5-M, SAMG lifts the overall score from 0.63 to 0.66, with noticeable gains in challenging categories like "Counting", "Position", and "Attribute Binding". This structural improvement suggests that our spatially adaptive guidance effectively mitigates semantic bleeding and attribute confusion, leading to highly accurate compositional synthesis.

\begin{table*}[t]
  \centering
  \caption{Ablation study on SD 1.5 evaluating the impact of global CFG scales, SAMG bounds $[\omega_{\min}, \omega_{\max}]$, and spatial smoothing (Kernel). The best results overall are highlighted in \textbf{bold}, and the best results within the CFG baseline are \underline{underlined}.}
  \label{tab:ablation}
  \resizebox{\textwidth}{!}{
  \begin{tabular}{lcc cccc ccc}
    \toprule
    \multirow{2}{*}{\textbf{Method}} & \multirow{2}{*}{\textbf{Scale / Bounds}} & \multirow{2}{*}{\textbf{Kernel}} & \multicolumn{7}{c}{\textbf{Metrics}} \\
    \cmidrule(lr){4-10}
    & & & HPSv2 $\uparrow$ & ImageReward $\uparrow$ & Aesthetic $\uparrow$ & CLIP $\uparrow$ & Top-1 $\uparrow$ & Top-5 $\uparrow$ & Top-10 $\uparrow$ \\
    \midrule
    CFG & 5.0 & - & 26.66 & -0.022 & 5.566 & 65.53 & 86 & 98 & 98 \\
    CFG & 10.0 & - & 26.75 & 0.015 & 5.612 & 65.73 & 88 & 98 & 98 \\
    CFG & 12.0 & - & 26.68 & -0.010 & 5.605 & \underline{66.25} & 87 & 97 & 98 \\
    CFG & 7.5 & - & \underline{26.88} & \underline{0.053} & \underline{5.633} & 66.08 & \underline{89} & \underline{99} & \underline{99} \\
    \midrule
    SAMG & [5, 10] & 3 & 26.91 & 0.045 & 5.645 & 66.42 & 89 & 99 & 99 \\
    SAMG & [5, 10] & 1 & 26.95 & 0.051 & 5.651 & 66.55 & 90 & 99 & 99 \\
    SAMG & [5, 12] & 3 & 26.98 & 0.059 & 5.660 & 66.68 & 91 & 99 & 99 \\
    SAMG & [5, 12] & 1 & \textbf{27.05} & \textbf{0.068} & \textbf{5.672} & \textbf{66.85} & \textbf{92} & \textbf{100} & \textbf{100} \\
    \bottomrule
  \end{tabular}
  }
\end{table*}

\subsection{Text-to-Video Generation}

To evaluate the generalization capability of our method across modalities and temporal extensions, we conducted a quantitative comparison on the ChronoMagic-Bench-150 dataset using two mainstream text-to-video generation models: CogVideoX-2B and ModelScope-1.7B. As shown in Table \ref{tab:chrono_eval}, SAMG consistently outperforms the standard CFG baseline across all evaluation metrics.

\paragraph{Semantic and Temporal Alignment.} SAMG achieves improvements in both CLIP SIM and MTScore CLIP, which measure the semantic alignment between the text prompt and the generated video. For instance, on ModelScope-1.7B, the MTScore CLIP increases from 10.56 to 10.78. This indicates that SAMG's dynamic energy modulation is effective not only for static images but also for accurately injecting semantic information in complex, temporally-extended alignment tasks.

\paragraph{Frame-level Visual Quality.} Standard CFG often triggers frame-wise color over-saturation or structural artifacts during video generation. SAMG successfully mitigates this phenomenon. Quantitatively, SAMG yields a significant reduction in Frame LPIPS (e.g., dropping from 8.12 to 7.35 on ModelScope-1.7B) while simultaneously improving Frame SSIM. This demonstrates that our conservative boundary strategy effectively preserves micro-textures and structural integrity within individual video frames.

\paragraph{Motion Smoothness.} In terms of CHScore Flow, which evaluates the naturalness and consistency of inter-frame motion, SAMG achieves higher scores on both CogVideoX-2B (60.15 vs. 59.93) and ModelScope-1.7B (72.10 vs. 70.47). This implies that SAMG effectively prevents the manifold deviation typically caused by a globally uniform high guidance scale, resulting in smoother and more coherent transitions between frames.

In summary, as a training-free, plug-and-play sampler, SAMG not only resolves the ``detail-artifact dilemma'' in the image domain but also demonstrates that its core geometric manifold constraint mechanism is highly effective and seamlessly transferable to high-dimensional video generation tasks.

\subsection{Ablation Studies}
\label{subsec:ablations}
We conduct ablation studies on the SD 1.5 architecture (Table \ref{tab:ablation}) to validate SAMG's components: global CFG scales, adaptive boundary parameters ($\omega_{\min}$ and $\omega_{\max}$), and pixel-wise independent mapping.

\paragraph{Impact of Global CFG Scales.} Evaluating global CFG from 5.0 to 12.0 confirms that CFG 7.5 is the optimal baseline. Forcing CFG to 12.0 slightly improves the CLIP score but significantly degrades human preference metrics (HPSv2 drops to 26.68, ImageReward to -0.010), leading to over-saturation and structural artifacts. This demonstrates that aggressively scaling standard CFG pushes latents off the natural image manifold.

\paragraph{Impact of Adaptive Bounds ($\omega_{\min}$ and $\omega_{\max}$).} We evaluate SAMG with scale ranges of $[5.0, 10.0]$ and $[5.0, 12.0]$, both outperforming global CFG baselines. The $[5.0, 12.0]$ configuration achieves the best balance: anchoring high-frequency boundaries at $\omega_{\min}=5.0$ preserves structural integrity, while allowing $\omega_{\max}=12.0$ in smooth regions maximizes semantic injection.

\paragraph{Preventing Energy Leakage (Kernel Size).} To validate the necessity of strict pixel-wise independence, we compare point-wise SAMG (\textbf{Kernel = 1}) against a spatially smoothed variant (\textbf{Kernel = 3}). As shown in Table \ref{tab:ablation}, spatial smoothing consistently degrades performance. Smoothing causes "energy leakage"—blurring conservative constraints into safe regions and aggressive guidance into delicate boundaries—confirming that pixel-wise mapping is essential for accurate manifold adherence. 

Collectively, these findings highlight the inherent limitations of rigid global scaling. They empirically validate that our geometrically-driven, spatially-aware guidance strategy is critical for maximizing text-image alignment without compromising the structural fidelity of the generated latents.




\section{Conclusion}
\label{sec:conclusion}
In this paper, we identified that the globally uniform scalar formulation of Classifier-Free Guidance intrinsically causes manifold deviation, trapping diffusion models in a detail-artifact dilemma. By viewing CFG through the lens of differential geometry, we established a theoretical upper bound for spatially adaptive guidance and proposed SAMG, a training-free, virtually zero-cost sampling algorithm. SAMG dynamically modulates guidance scales based on local conditional energy, preserving delicate micro-textures and ensuring temporal consistency in videos while maximizing overall semantic alignment. Extensive empirical evaluations across both static image and video generation tasks confirm its superiority. Furthermore, by constructing geometrically sound target manifolds, integrating SAMG into distillation pipelines could fundamentally elevate the theoretical upper bound of modern CFG-free models that currently inherit teacher artifacts.

\appendix

\section{Theoretical Extensions}
\label{sec:appendix_proofs}

In this appendix, we provide the rigorous mathematical derivations and proofs for the theoretical claims presented in the main text. Specifically, we explicitly formalize the generative sampling trajectory on a Riemannian manifold, rigorously prove the manifold deviation bound formulated in Proposition 1, establish the geometric safety of the first-order Taylor relaxation via strict convexity, and theoretically generalize our geometric constraints to continuous-time Flow Matching ordinary differential equations (ODEs).

\subsection{Geometric Setup}
\label{subsec:app_geometric_setup}

Let the natural visual data distribution be supported on a smooth Riemannian manifold $\mathcal{M} \subset \mathbb{R}^D$ ($D = C \times H \times W$), equipped with the induced Euclidean metric. 

During the reverse process at timestep $t$, the unconditional score prediction implicitly projects the noisy state $\mathbf{z}_t$ onto $\mathcal{M}$. We define this unguided anchor point $\mathbf{p} \in \mathcal{M}$ as:
\begin{equation}
    \mathbf{p} \triangleq \frac{\mathbf{z}_t - \sqrt{1-\bar{\alpha}_t} \epsilon_u}{\sqrt{\bar{\alpha}_t}}
\end{equation}

Classifier-Free Guidance (CFG) perturbs this anchor along the conditional direction. Let the Delta Score be $\Delta \epsilon_t \triangleq \epsilon_c - \epsilon_u$. We define the normalized unit tangent vector $\mathbf{v}(x) \in T_{\mathbf{p}}\mathcal{M}$ and the local conditional energy $E_t(x)$ at spatial coordinate $x$ as:
\begin{equation}
    \mathbf{v}(x) \triangleq -\frac{\Delta \epsilon_t(x)}{\|\Delta \epsilon_t(x)\|_2}, \quad E_t(x) \triangleq \frac{1}{C}\|\Delta \epsilon_t(x)\|_2^2
\end{equation}

The CFG operation structurally corresponds to a linear extrapolation in the ambient space $\mathbb{R}^D$. The explicit coordinate $y(s)$ and the step size $s(x)$ evaluated along the tangent space are formulated as:
\begin{equation}
    y(s) = \mathbf{p} + s(x) \mathbf{v}(x)
\end{equation}
\begin{equation}
\label{eq:app_step_size}
    s(x) = \left\| \omega \frac{\sqrt{1-\bar{\alpha}_t}}{\sqrt{\bar{\alpha}_t}} \Delta \epsilon_t(x) \right\|_2 = \omega \sqrt{c_t E_t(x)}
\end{equation}
where $c_t \triangleq C \frac{1-\bar{\alpha}_t}{\bar{\alpha}_t}$ is a deterministic scalar dependent on the forward schedule.

\subsection{Proof of Proposition 1}
\label{subsec:app_proof_prop1}

Because $\mathcal{M}$ is highly non-linear, a linear extrapolation $y(s)$ along $T_{\mathbf{p}}\mathcal{M}$ inherently introduces an orthogonal deviation from the true manifold. The geometrically optimal trajectory bounded on $\mathcal{M}$ is the Riemannian geodesic $\gamma: \mathbb{R} \to \mathcal{M}$, given by the exponential map $\gamma(s) = \exp_{\mathbf{p}}(s \mathbf{v}(x))$.

\vspace{0.5em}
\noindent \textbf{Proposition 1.} \textit{For a sufficiently small step size $s$, let $\kappa(x)$ be the normal curvature along $\mathbf{v}(x)$. The deviation error $\mathcal{E}_{dev}(x) = \| y - \gamma(s) \|_2$ is bounded by $\mathcal{E}_{dev}(x) = \frac{1}{2} \kappa(x) s^2 + \mathcal{O}(s^3)$. To restrict the generation trajectory within a $\delta$-neighborhood ($\mathcal{E}_{dev}(x) \le \delta$), the optimal guidance scale must satisfy $\omega_{ideal}(x) \le \sqrt{\frac{2\delta}{\kappa(x) c_t E_t(x)}}$.}

\begin{proof}
By the definition of a geodesic, $\gamma(s)$ satisfies the initial conditions $\gamma(0) = \mathbf{p}$ and $\dot{\gamma}(0) = \mathbf{v}(x)$. The second-order Taylor expansion of $\gamma(s)$ in the ambient space $\mathbb{R}^D$ is:
\begin{equation}
    \gamma(s) = \gamma(0) + s \dot{\gamma}(0) + \frac{s^2}{2} \ddot{\gamma}(0) + \mathcal{O}(s^3)
\end{equation}

Since $\gamma(s)$ is a geodesic, its intrinsic acceleration vanishes ($\nabla_{\dot{\gamma}}\dot{\gamma} = 0$). By the Gauss Formula for embedded sub-manifolds, the ambient acceleration $\ddot{\gamma}(0)$ is strictly orthogonal to $T_{\mathbf{p}}\mathcal{M}$ and is governed entirely by the Second Fundamental Form $\mathrm{II}$:
\begin{equation}
    \ddot{\gamma}(0) = \mathrm{II}(\dot{\gamma}(0), \dot{\gamma}(0)) = \mathrm{II}(\mathbf{v}(x), \mathbf{v}(x))
\end{equation}

The orthogonal error vector $e(s)$ between the CFG linear step $y(s)$ and the true manifold curve $\gamma(s)$ is:
\begin{align}
    e(s) &= y(s) - \gamma(s) \nonumber \\
         &= \big(\mathbf{p} + s \mathbf{v}(x)\big) - \left( \mathbf{p} + s \mathbf{v}(x) + \frac{s^2}{2} \mathrm{II}(\mathbf{v}(x), \mathbf{v}(x)) + \mathcal{O}(s^3) \right) \nonumber \\
         &= -\frac{s^2}{2} \mathrm{II}(\mathbf{v}(x), \mathbf{v}(x)) - \mathcal{O}(s^3)
\end{align}

Let $\kappa(x) \triangleq \|\mathrm{II}(\mathbf{v}(x), \mathbf{v}(x))\|_2$ denote the normal curvature. The manifold deviation error simplifies to:
\begin{equation}
    \mathcal{E}_{dev}(x) = \| e(s) \|_2 = \frac{1}{2} \kappa(x) s^2 + \mathcal{O}(s^3) \approx \frac{1}{2} \kappa(x) s^2
\end{equation}

Substituting the explicit step size $s^2 = \omega^2 c_t E_t(x)$ derived from Equation (\ref{eq:app_step_size}):
\begin{equation}
    \mathcal{E}_{dev}(x) \approx \frac{1}{2} \kappa(x) \omega^2 c_t E_t(x)
\end{equation}

Enforcing the geometric bounding condition $\mathcal{E}_{dev}(x) \le \delta$ yields:
\begin{equation}
    \omega^2 \le \frac{2\delta}{\kappa(x) c_t E_t(x)} \implies \omega_{ideal}(x) \le \sqrt{\frac{2\delta}{\kappa(x) c_t E_t(x)}}
\end{equation}
\end{proof}

\subsection{Conservative Relaxation Bound}
\label{subsec:app_convexity_proof}

In practice, computing the exact high-dimensional curvature $\kappa(x)$ is intractable. We relax the theoretical bound $\omega \propto E_t(x)^{-1/2}$ via a first-order Taylor expansion. We formalize that this linear relaxation constitutes a strict mathematical lower bound, ensuring geometric safety.

\vspace{0.5em}
\noindent \textbf{Supplementary Theorem 1.} \textit{Let $f(E) = E^{-1/2}$ represent the target boundary constraint for $E \in \mathbb{R}_{>0}$. Its first-order Taylor expansion $g(E)$ at any operating point $\eta_0 > 0$ strictly lower-bounds $f(E)$ (i.e., $g(E) < f(E)$ for all $E \neq \eta_0$).}

\begin{proof}
The first and second derivatives of $f(E)$ are:
\begin{equation}
    f'(E) = -\frac{1}{2} E^{-3/2}, \quad f''(E) = \frac{3}{4} E^{-5/2}
\end{equation}
For any valid guidance energy $E > 0$, we have $f''(E) > 0$, confirming that $f(E)$ is strictly convex on $\mathbb{R}_{>0}$. 

By Taylor's Theorem with the Lagrange remainder, for any $E > 0$ and $E \neq \eta_0$, there exists a $\xi$ between $\eta_0$ and $E$ such that:
\begin{equation}
    f(E) = f(\eta_0) + f'(\eta_0)(E - \eta_0) + \frac{f''(\xi)}{2}(E - \eta_0)^2
\end{equation}

We define the first-order approximation $g(E)$ as the tangent line at $\eta_0$:
\begin{equation}
    g(E) \triangleq f(\eta_0) + f'(\eta_0)(E - \eta_0) = \underbrace{\frac{3}{2}\eta_0^{-1/2}}_{C_1 > 0} - \underbrace{\frac{1}{2}\eta_0^{-3/2}}_{C_2 > 0} E
\end{equation}

Since $f''(\xi) > 0$ and $(E - \eta_0)^2 > 0$, the remainder term is strictly positive:
\begin{equation}
    \frac{f''(\xi)}{2}(E - \eta_0)^2 > 0
\end{equation}

Therefore, it strictly holds that:
\begin{equation}
    f(E) = g(E) + \frac{f''(\xi)}{2}(E - \eta_0)^2 \implies g(E) < f(E) \quad \forall E \neq \eta_0
\end{equation}

This proves that the negative-slope linear function $g(E) = C_1 - C_2 E$ mathematically guarantees under-approximation of $f(E)$. The practical SAMG affine mapping $\boldsymbol{\Omega}_{map}(x)$ instantiates this $g(E)$ structure, providing a conservative lower bound that structurally prevents manifold over-extrapolation.
\end{proof}

\subsection{Generalization to Continuous-Time Flow Matching ODEs}
\label{subsec:app_flow_matching}

The geometric constraints rigorously generalize from discrete Tweedie projections to continuous-time Flow Matching and Rectified Flow frameworks. Let $\mathbf{z}_t$ follow the exact probability flow ODE governed by the modified velocity field $\tilde{\mathbf{v}}_\omega$:
\begin{equation}
    \frac{\mathrm{d}\mathbf{z}_t}{\mathrm{d}t} = \tilde{\mathbf{v}}_\omega(\mathbf{z}_t, t) = \mathbf{v}_u(\mathbf{z}_t, t) + \omega \Delta \mathbf{v}(\mathbf{z}_t, t)
\end{equation}
where $\Delta \mathbf{v} \triangleq \mathbf{v}_c - \mathbf{v}_u$. The numerical Euler integration step is explicitly given by $y = \mathbf{z}_t - \Delta t \tilde{\mathbf{v}}_\omega$.

\vspace{0.5em}
\noindent \textbf{Theorem A.2 (ODE Local Truncation Deviation).} \textit{Let $\Psi_{t \to t-\Delta t}(\mathbf{z}_t)$ denote the exact integral curve of the vector field $\tilde{\mathbf{v}}_\omega$. Let $E_{\mathbf{v}}(x) = \frac{1}{C}\|\Delta \mathbf{v}(x)\|_2^2$ be the flow guidance energy. The local orthogonal manifold deviation $\mathcal{E}_{ODE} = \|\Psi_{t \to t-\Delta t}(\mathbf{z}_t) - y\|_2$ requires the continuous-time guidance scale to be bounded by:}
\begin{equation}
    \omega(x) \le \frac{1}{\Delta t} \sqrt{\frac{2\delta}{\kappa_{\mathcal{F}}(x) C E_{\mathbf{v}}(x)}}
\end{equation}
\textit{where $\kappa_{\mathcal{F}}(x)$ is the path curvature of the conditional vector field.}

\begin{proof}
By the Picard-Lindelöf theorem, the true solution admits the Taylor expansion:
\begin{equation}
    \Psi_{t \to t-\Delta t}(\mathbf{z}_t) = \mathbf{z}_t - \Delta t \tilde{\mathbf{v}}_\omega + \frac{(\Delta t)^2}{2} \frac{\mathrm{d}\tilde{\mathbf{v}}_\omega}{\mathrm{d}t} - \mathcal{O}\big((\Delta t)^3\big)
\end{equation}
The local truncation error reduces to the acceleration norm orthogonal to the flow:
\begin{equation}
    \mathcal{E}_{ODE} = \frac{(\Delta t)^2}{2} \left\| \frac{\partial \tilde{\mathbf{v}}_\omega}{\partial t} + \big(\nabla_{\mathbf{z}_t} \tilde{\mathbf{v}}_\omega\big) \tilde{\mathbf{v}}_\omega \right\|_2 + \mathcal{O}\big((\Delta t)^3\big)
\end{equation}
Isolating the dominating quadratic guidance component for $\omega \gg 1$:
\begin{equation}
    \mathcal{E}_{ODE} \approx \frac{(\Delta t)^2}{2} \omega^2 \left\| \big(\nabla_{\mathbf{z}_t} \Delta \mathbf{v}\big) \Delta \mathbf{v} \right\|_2 \le \frac{1}{2} \kappa_{\mathcal{F}}(x) \omega^2 (\Delta t)^2 \|\Delta \mathbf{v}\|_2^2
\end{equation}
Substituting $\|\Delta \mathbf{v}(x)\|_2^2 = C E_{\mathbf{v}}(x)$ and strictly enforcing the geometric safety condition $\mathcal{E}_{ODE} \le \delta$ yields:
\begin{equation}
    \frac{1}{2} \kappa_{\mathcal{F}}(x) \omega^2 (\Delta t)^2 C E_{\mathbf{v}}(x) \le \delta \implies \omega(x) \le \sqrt{\frac{2\delta}{\kappa_{\mathcal{F}}(x) \underbrace{C(\Delta t)^2}_{c_t} E_{\mathbf{v}}(x)}}
\end{equation}
This confirms mathematical isomorphism to the discrete diffusion curvature bound under the time discretization mapping $c_t = C(\Delta t)^2$.
\end{proof}

\subsection{Global Trajectory Deviation Accumulation Bound}
\label{subsec:app_global_bound}

We formalize how local manifold deviation errors iteratively accumulate over $N$ generative sampling steps to govern global structural integrity.

\vspace{0.5em}
\noindent \textbf{Theorem A.3 (Global Gr\"onwall Error Propagation).} \textit{Let $\mathbf{z}_k$ denote the generative state at step $k$. Assume the discrete update function $F_k(\mathbf{z}) = \mathbf{z} - h \tilde{\epsilon}_\omega(\mathbf{z})$ is globally $L$-Lipschitz. If the local single-step orthogonal deviation is rigidly constrained point-wise such that $\mathcal{E}_{dev}^{(k)}(x) \le \delta$, the cumulative manifold deviation $\mathcal{E}_{global}^{(N)}$ after $N$ steps is bounded by:}
\begin{equation}
    \mathcal{E}_{global}^{(N)} \le \frac{\delta}{L h} \Big( \exp(L N h) - 1 \Big)
\end{equation}

\begin{proof}
Let $\Pi_{\mathcal{M}_k}(\mathbf{z}_k) = \arg\min_{\mathbf{y} \in \mathcal{M}_k} \|\mathbf{z}_k - \mathbf{y}\|_2$ denote the exact projection onto the true data sub-manifold $\mathcal{M}_k$. The propagated error is $e_{k} = \|\mathbf{z}_{k} - \Pi_{\mathcal{M}_{k}}(\mathbf{z}_{k})\|_2$. By the triangle inequality and the Lipschitz condition:
\begin{align}
    e_{k} &\le \left\| F_{k-1}(\mathbf{z}_{k-1}) - F_{k-1}(\Pi_{\mathcal{M}_{k-1}}(\mathbf{z}_{k-1})) \right\|_2 + \mathcal{E}_{dev}^{(k-1)} \nonumber \\
          &\le (1 + L h) \|\mathbf{z}_{k-1} - \Pi_{\mathcal{M}_{k-1}}(\mathbf{z}_{k-1})\|_2 + \delta \nonumber \\
          &= (1 + L h) e_{k-1} + \delta
\end{align}
Unrolling this recursive inequality from $e_0 = 0$ (perfect initialization on the Gaussian prior):
\begin{equation}
    e_N \le \delta \sum_{i=0}^{N-1} (1 + Lh)^i = \delta \frac{(1 + Lh)^N - 1}{(1 + Lh) - 1} = \frac{\delta}{Lh} \Big( (1 + Lh)^N - 1 \Big)
\end{equation}
Applying the classical limit inequality $(1 + \frac{x}{n})^n \le \exp(x)$ for strictly positive variables:
\begin{equation}
    \mathcal{E}_{global}^{(N)} = e_N \le \frac{\delta}{Lh} \Big( \exp(L N h) - 1 \Big)
\end{equation}
This proves that homogeneously applying an unbounded global $\omega$ induces an exponential global manifold divergence, fundamentally establishing the necessity for adaptive energy bounds.
\end{proof}

\subsection{Curvature Bound via Spectral Norm of Score Jacobian}
\label{subsec:app_score_jacobian}

The true Riemannian normal curvature $\kappa(x)$ relies on the intractable Second Fundamental Form. We strictly upper-bound this intrinsic curvature utilizing the spectral properties of the continuous Score Jacobian.

\vspace{0.5em}
\noindent \textbf{Lemma A.4 (Score-Hessian Isomorphism).} \textit{Let $\mathcal{H}_t(\mathbf{z}_t) \\ = \nabla_{\mathbf{z}_t}^2 \log p_t(\mathbf{z}_t)$ represent the exact Hessian of the log-density. Under Tweedie's parameterization, the Hessian is isomorphic to the Score Jacobian $\mathbf{J}_{\epsilon_u}$:}
\begin{equation}
    \mathcal{H}_t(\mathbf{z}_t) = -\frac{1}{\sqrt{1-\bar{\alpha}_t}} \mathbf{J}_{\epsilon_u}(\mathbf{z}_t)
\end{equation}

\vspace{0.5em}
\noindent \textbf{Theorem A.5 (Spectral Curvature Bound).} \textit{The extrinsic normal curvature $\kappa(x)$ evaluated along the guidance vector $\mathbf{v}(x) = -{\Delta \epsilon_t}/{\|\Delta \epsilon_t\|_2}$ is strictly bounded by the local spectral norm $\rho(\cdot)$:}
\begin{equation}
    \kappa(x) \le \frac{\rho\big(\mathbf{J}_{\epsilon_u}(x)\big)}{\|\epsilon_u(x)\|_2}
\end{equation}

\begin{proof}
By the differential geometry of implicit surfaces, the normal curvature along a unit tangent $\mathbf{v} \in T_x\mathcal{M}$ of the probability level set $\log p_t(x) = C$ is geometrically defined by:
\begin{equation}
    \kappa(x) = \frac{|\mathbf{v}^T \mathcal{H}_t(x) \mathbf{v}|}{\|\nabla_{\mathbf{z}_t} \log p_t(x)\|_2} 
\end{equation}
Applying the Rayleigh quotient property for the symmetric matrix $\mathcal{H}_t(x)$:
\begin{equation}
    \left| \mathbf{v}^T \mathcal{H}_t(x) \mathbf{v} \right| \le \| \mathcal{H}_t(x) \|_2 = \rho\big(\mathcal{H}_t(x)\big)
\end{equation}
Substituting Lemma A.4 and the exact continuous correspondence $\|\nabla_{\mathbf{z}_t} \log p_t(x)\|_2 = \|\epsilon_u(x)\|_2 / \sqrt{1-\bar{\alpha}_t}$ yields the supremum boundary condition:
\begin{equation}
    \kappa(x) \le \frac{1}{\|\nabla_{\mathbf{z}_t} \log p_t(x)\|_2} \left( \frac{1}{\sqrt{1-\bar{\alpha}_t}} \rho\big(\mathbf{J}_{\epsilon_u}(x)\big) \right) = \frac{\rho\big(\mathbf{J}_{\epsilon_u}(x)\big)}{\|\epsilon_u(x)\|_2}
\end{equation}
This structurally bridges the manifold geometry with network parameterization, proving that delicate boundary regions exhibiting high curvature are precisely governed by large local score gradients.
\end{proof}

\subsection{Sub-Optimality of Spatial Smoothing via Jensen's Inequality}
\label{subsec:app_jensen}

We theoretically formulate the detrimental geometric penalty of applying spatial convolutional smoothing to the energy map (formally validating the empirical necessity of \textbf{Kernel = 1} in Section 4.4).

\vspace{0.5em}
\noindent \textbf{Theorem A.6 (Smoothing Manifold Violation).} \textit{Let $\mathcal{K}_\sigma(u)$ be a normalized convolution kernel satisfying $\int \mathcal{K}_\sigma(u)du = 1$. The smoothed guidance energy is $\bar{E}_t = \mathcal{K}_\sigma * E_t$. For the strict convex boundary function $f(E) = E^{-1/2}$, spatial smoothing mathematically forces the applied guidance scale to violate the theoretically safe manifold bound at local high-frequency coordinates.}

\begin{proof}
Because $f''(E) = \frac{3}{4} E^{-5/2} > 0$ for all valid $E > 0$, the constraint function $f(E)$ is strictly convex. Applying Jensen's Inequality to the continuous spatial convolution integral:
\begin{equation}
    f\big(\bar{E}_t(x)\big) = f\left( \int_{\Omega} \mathcal{K}_\sigma(u) E_t(x - u) du \right) \le \int_{\Omega} \mathcal{K}_\sigma(u) f\big(E_t(x - u)\big) du
\end{equation}
Consider a high-frequency spatial boundary coordinate $x_b$, defined as a strict local maximum of the energy field: $E_t(x_b) > E_t(x_b - u)$ for all $u \neq 0$ within the kernel support. The local weighted average inherently attenuates this maximum:
\begin{equation}
    \bar{E}_t(x_b) = \int_{\Omega} \mathcal{K}_\sigma(u) E_t(x_b - u) du < \int_{\Omega} \mathcal{K}_\sigma(u) E_t(x_b) du = E_t(x_b)
\end{equation}
Since $f(E)$ is strictly monotonically decreasing, the attenuated smoothed evaluation inevitably yields an over-extrapolated scale coefficient:
\begin{equation}
    \bar{E}_t(x_b) < E_t(x_b) \implies f\big(\bar{E}_t(x_b)\big) > f\big(E_t(x_b)\big)
\end{equation}
Therefore, deriving the practical scale $\bar{\omega}(x_b)$ via the smoothed energy rigorously violates the upper limit $\omega_{ideal}$ derived in Proposition 1:
\begin{equation}
    \bar{\omega}(x_b) \propto \bar{E}_t(x_b)^{-1/2} > E_t(x_b)^{-1/2} \propto \omega_{ideal}(x_b)
\end{equation}
\begin{equation}
    \bar{\omega}(x_b) > \omega_{ideal}(x_b) \implies \bar{\mathcal{E}}_{dev}(x_b) > \delta
\end{equation}
This mathematically proves that point-wise independent calculations (where $\mathcal{K}_0$ is the Dirac delta distribution) are theoretically mandated. Spatial smoothing inherently induces geometric energy leakage, forcing orthogonal manifold over-extrapolation exactly at delicate structural boundaries.
\end{proof}

\section{Related Work}
\label{sec:related_work}

\subsection{Diffusion Models and Classifier-Free Guidance}
Generative visual modeling has been fundamentally transformed by score-based diffusion processes~\cite{ho2020denoisingdiffusionprobabilisticmodels, song2021scorebasedgenerativemodelingstochastic}. To alleviate the immense computational burden of pixel-space modeling, Latent Diffusion Models (LDMs)~\cite{rombach2022highresolutionimagesynthesislatent} introduced compressed autoencoding, establishing the foundation for high-resolution text-to-image architectures like Stable Diffusion 1.5 and SDXL~\cite{podell2023sdxlimprovinglatentdiffusion}. More recently, the field has increasingly converged towards continuous-time Flow Matching~\cite{lipman2023flowmatchinggenerativemodeling} and Rectified Flow~\cite{liu2022flowstraightfastlearning} paradigms. These mathematical frameworks govern state-of-the-art flow-matched Diffusion Transformers like SD3.5~\cite{esser2024scalingrectifiedflowtransformers}, and smoothly extend to complex text-to-video (T2V) architectures such as CogVideoX~\cite{yang2024cogvideox}. 

Despite these architectural advancements, aligning the generated outputs with textual prompts critically depends on conditioning mechanisms. Classifier-Free Guidance (CFG)~\cite{ho2022classifierfreediffusionguidance} is the ubiquitous standard. By linearly extrapolating the generative vector field towards the conditional estimate, CFG effectively acts as an implicit classifier gradient. However, its reliance on a rigid, globally uniform scalar locks the generation into a strict trade-off: insufficient scales fail to inject complex semantics, while high scales force the generative trajectory off the natural data manifold, triggering severe over-saturation and structural collapse~\cite{saharia2022photorealistictexttoimagediffusionmodels}.

\begin{figure*}[t]
  \centering
  \includegraphics[width=\linewidth]{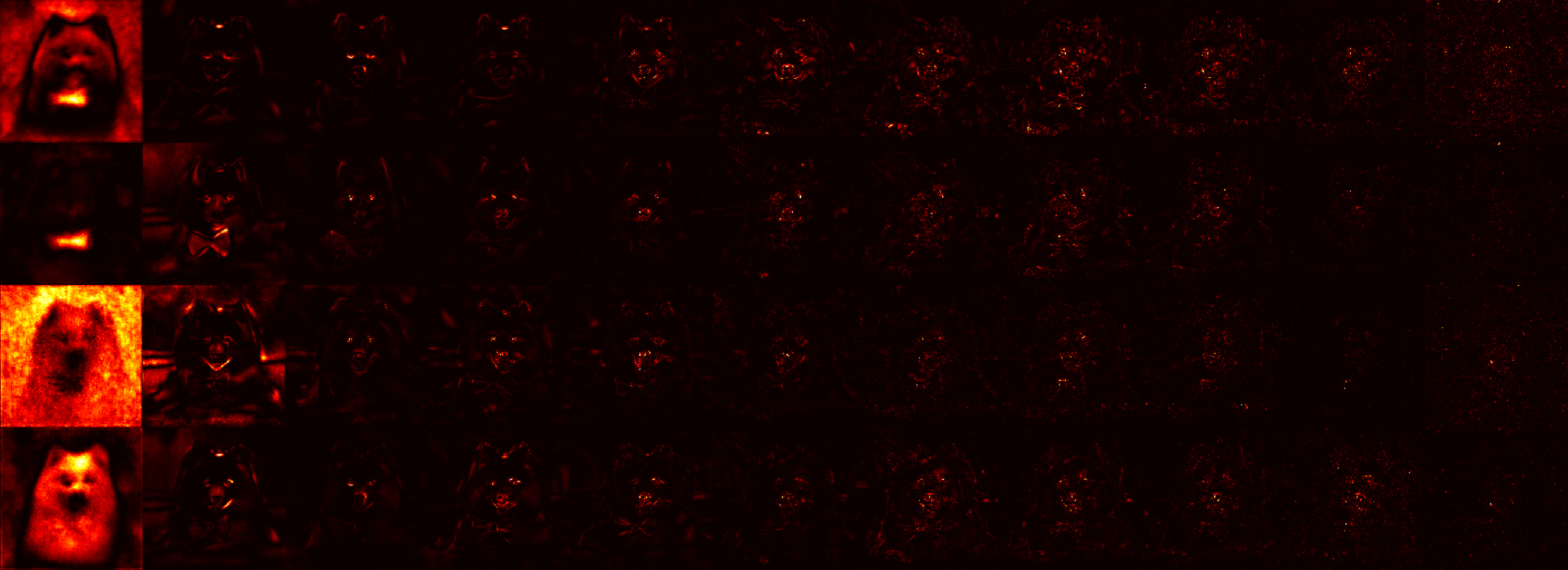}
 \caption{Visualization of decoupled latent energy evolution across the four SDXL channels. 
\textbf{Top to Bottom:} Ch 0 (Luma/Structure), Ch 1 (Warm/Cool), Ch 2 (Green/Magenta), and Ch 3 (High-Frequency). 
Brightness represents the local Delta Score magnitude. 
The visualization reveals a clear \textit{functional decoupling}: the structural and high-frequency channels (Ch 0, 3) exhibit sharper, more concentrated energy peaks during early manifold formation, while color channels (Ch 1, 2) show smoother, delayed stabilization. 
This spatial-temporal asymmetry across channels provides empirical evidence that uniform guidance is sub-optimal across feature hierarchies, supporting the potential for channel-wise adaptive modulation.}
  \label{fig:energy_montage}
\end{figure*}

\subsection{Mitigating the Detail-Artifact Dilemma}
To alleviate the artifacts induced by high global guidance scales, numerous training-free strategies have been proposed. Value-clipping methods, such as Dynamic Thresholding~\cite{saharia2022photorealistictexttoimagediffusionmodels} and CFG Rescale~\cite{lin2024commondiffusionnoiseschedules}, attempt to normalize latent variance during inference to prevent over-exposure. Other approaches fundamentally intervene in the network's feature processing: FreeU~\cite{si2023freeufreelunchdiffusion} re-weights the high- and low-frequency components within U-Net skip connections to balance semantics and structural details, while PAG~\cite{ahn2024self} introduces an attention-perturbed self-guidance path. Recently, geometrically motivated approaches like CFG++~\cite{chung2024cfgmanifoldconstrainedclassifierfree} and CFG-Zero~\cite{fan2025cfgzerostar} aim to regularize the tangential projection to remain bounded to the unconditional manifold. 

While these methods successfully mitigate global artifacts, they apply homogeneous temporal schedules or global feature modifications, completely ignoring the highly non-uniform spatial complexity of visual data. Recognizing that distinct regions demand varying levels of semantic injection, recent works have explored spatial guidance variations. For instance, Semantic-aware CFG (S-CFG)~\cite{shen2024rethinkingspatialinconsistencyclassifierfree} utilizes cross-attention maps to assign customized CFG scales to different semantic units. However, relying on the heuristic extraction of high-dimensional attention tensors incurs significant computational overhead. Furthermore, they are highly architecture-specific, making them challenging to deploy out-of-the-box on modern Joint Multi-Modal DiT architectures (e.g., SD3.5) where cross-attention and self-attention are intricately fused. In contrast, SAMG operates strictly on the output delta score ($\Delta \epsilon_t$), achieving spatially adaptive modulation with zero computational overhead and offering universal plug-and-play compatibility.

\subsection{Geometric Perspectives on Generative Manifolds}
Our methodology is deeply rooted in the geometric interpretation of generative models. Recent studies conceptualize the diffusion reverse process as evolving along a highly non-linear, curved Riemannian manifold embedded within a high-dimensional space~\cite{karras2022elucidatingdesignspacediffusionbased, debortoli2022riemannianscorebasedgenerativemodelling}. Theoretical analyses have demonstrated that employing linear tangential extrapolation in such curved spaces inherently introduces orthogonal deviation errors~\cite{chung2024cfgmanifoldconstrainedclassifierfree, bao2022analyticdpmanalyticestimateoptimal}. Furthermore, through the lens of Tweedie's Formula~\cite{Good1992}, score networks implicitly project noisy states back onto the local data sub-manifold. While previous works (e.g., CFG++) largely rely on this manifold deviation at a global level to justify modifying the unconditional anchor, SAMG takes a critical step forward. We explicitly formulate an adaptive spatial bounding mechanism based on pixel-wise curvature approximation. By translating intractable high-dimensional geometric constraints into a practical, negative-slope energy mapping, SAMG physically prevents orthogonal manifold deviation exactly at structurally delicate boundaries.

\section{Latent Channel-wise Energy Evolution}
\label{sec:appendix_channels}

To further investigate the fine-grained geometric behavior of SAMG, we decouple the guidance energy $\Delta \epsilon_t$ into its four constituent latent channels as defined in the SDXL VAE space. Figure~\ref{fig:energy_montage} presents a seamless montage of these decoupled energy maps across the denoising trajectory, visualized using a thermal colormap to highlight the intensity of the Delta Score.

As illustrated, the energy distribution exhibits significant functional decoupling across channels. The \textbf{Structure Channel (Ch 0)} and \textbf{High-Frequency Channel (Ch 3)} dominate the early stages of manifold formation, where the model establishes global semantic layout and intricate micro-textures (e.g., fur patterns and sharp contours). In contrast, the \textbf{Color Channels (Ch 1, 2)} exhibit relatively smoother energy fields, primarily stabilizing in the intermediate steps to refine chromatic boundaries. 

This inherent asymmetry reinforces our core hypothesis: a globally uniform CFG scale is sub-optimal not only spatially but also across feature dimensions. Specifically, high-frequency channels exhibiting sharp energy peaks are more susceptible to orthogonal manifold deviation, thus requiring more conservative bounding. 

\paragraph{Future Outlook: Channel-wise Adaptive Modulation.} 
These observations pave the way for a more granular extension of our work—\textit{Channel-wise SAMG}. By dynamically modulating the guidance scale $\omega$ independently for each latent channel based on its specific frequency characteristics and curvature sensitivity, we could theoretically decouple structural preservation from semantic color injection. This would allow for an even more aggressive guidance on semantic-rich low-frequency components while maintaining absolute geometric integrity for delicate high-frequency details, further pushing the Pareto frontier of diffusion sampling.

\section{Limitations}
\label{sec:limitations}

\begin{figure}[t]
  \centering
  \includegraphics[width=\linewidth]{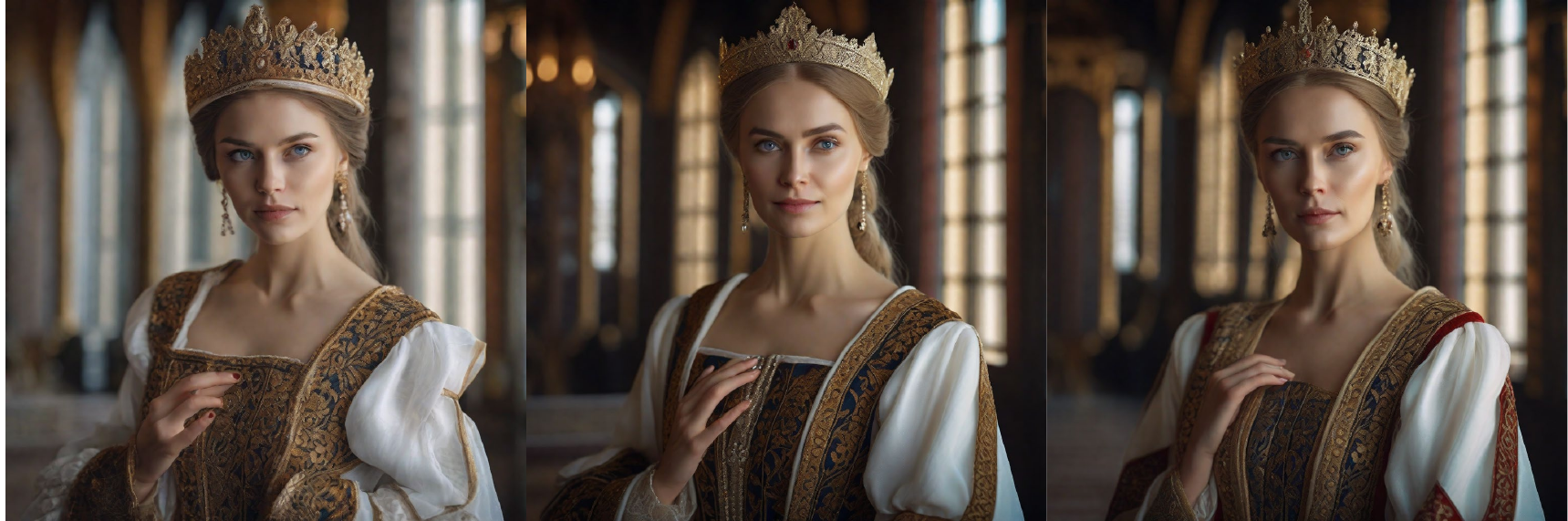}
  \caption{A limitation of SAMG in handling dense semantic overlaps. \textbf{Left:} Low Guidance. \textbf{Middle:} High Guidance. \textbf{Right:} SAMG. In regions where multiple intricate structures overlap (e.g., the earring and the hair), the latent energy merges into a single high-energy cluster. SAMG conservatively applies a low guidance scale, which prevents the severe artifacts of High CFG, but struggles to semantically disentangle the earring from the hair strands, resulting in a fused texture.}
  \label{fig:failure_case}
\end{figure}

While Spatial Adaptive Multi Guidance (SAMG) effectively resolves the detail-artifact dilemma in most scenarios, it encounters challenges in regions with dense semantic overlaps. Figure~\ref{fig:failure_case} illustrates a typical failure case involving overlapping micro-structures---specifically, an intricate earring resting against highly detailed hair.

Because SAMG modulates the guidance scale based on the spatial energy $E_t(x)$ evaluated in the highly compressed latent space (e.g., $128 \times 128$ for SDXL), it inherently faces a resolution bottleneck. When two distinct semantic entities with high-frequency details overlap spatially, they effectively merge into a single high-energy cluster within the local latent representation. Consequently, SAMG conservatively applies a minimum guidance scale ($\omega_{\min}$) to this entire cluster to prevent structural collapse. 

While this defensive mechanism successfully avoids the severe color over-saturation and structural degradation seen in standard High CFG (Figure~\ref{fig:failure_case}, Middle), it simultaneously fails to provide sufficient semantic push to distinctly disentangle the two objects. As a result, in the SAMG output (Figure~\ref{fig:failure_case}, Right), the earring appears structurally fused with the surrounding hair strands. Future work could address this limitation by incorporating cross-attention priors to semantically disentangle overlapping high-energy regions before applying adaptive scaling.

\section{Detailed Experimental Setup}
\label{sec:appendix_exp_setup}

In this section, we provide detailed descriptions of the diffusion models, guidance strategies, benchmarks, and evaluation metrics utilized in our experiments.

\subsection{Diffusion Models}

\textbf{Stable Diffusion 1.5 (SD 1.5)}~\cite{rombach2022highresolutionimagesynthesislatent}: A foundational text-to-image latent diffusion model that operates within a compressed autoencoded latent space. It employs a U-Net architecture to iteratively denoise latents, balancing generation fidelity with computational efficiency.

\textbf{Stable Diffusion XL (SDXL)}~\cite{podell2023sdxlimprovinglatentdiffusion}: An advanced iteration of the latent diffusion framework featuring a significantly expanded U-Net backbone and a dual text-encoder system. SDXL is designed for high-resolution synthesis and superior complex prompt comprehension.

\textbf{Stable Diffusion 3.5 Medium (SD3.5-M)}~\cite{esser2024scalingrectifiedflowtransformers}: A state-of-the-art continuous-time generative model based on the Rectified Flow matching paradigm. It utilizes a Joint Multi-Modal Diffusion Transformer architecture to transport probability mass along linear trajectories, offering enhanced semantic alignment and structural robustness.

\textbf{CogVideoX-2B}~\cite{yang2024cogvideox}: A large-scale text-to-video generation model that leverages a 3D causal Variational Autoencoder and an expert transformer architecture. It is specifically optimized for synthesizing temporally coherent and high-fidelity video sequences driven by textual prompts.

\textbf{ModelScope-1.7B}~\cite{wang2023modelscope}: A robust video diffusion model engineered to generate complex spatio-temporal dynamics and realistic motion patterns, demonstrating strong capabilities in prompt-driven video synthesis.

\subsection{Guidance Strategies}

\textbf{Classifier-Free Guidance (CFG)}~\cite{ho2022classifierfreediffusionguidance}: The ubiquitous standard for text-conditional generation. It modifies the predicted vector field by linearly extrapolating away from the unconditional prior towards the text-driven estimate, amplifying semantic alignment at the risk of structural deviation.

\textbf{Perturbed Attention Guidance (PAG)}~\cite{ahn2024self}: A guidance mechanism that generates a degraded unconditional path by systematically perturbing the self-attention maps during the diffusion process. This provides structural guidance without requiring explicit textual null-conditions.

\textbf{CFG++}~\cite{chung2024cfgmanifoldconstrainedclassifierfree}: A geometrically motivated variant of CFG that constrains the guidance update to remain on the unconditional data manifold, aiming to mitigate the over-saturation artifacts caused by standard tangential extrapolation.

\textbf{CFG-Zero}~\cite{fan2025cfgzerostar}: A recent guidance strategy that dynamically adjusts the unconditional prediction during the reverse process, enhancing semantic alignment and preserving sample quality without the need for additional manifold training.

\subsection{Benchmarks and Datasets}

\textbf{Pick-a-Pic}~\cite{kirstain2023pickapicopendatasetuser}: A large-scale dataset of text-to-image prompts collected from real users, which is highly effective for evaluating human preference and diverse generation scenarios.

\textbf{DrawBench}~\cite{saharia2022photorealistictexttoimagediffusionmodels}: A challenging prompt suite designed to test the fundamental capabilities of text-to-image models, including compositionality, spatial relations, and complex text comprehension.

\textbf{GenEval}~\cite{ghosh2023genevalobjectfocusedframeworkevaluating}: An objective, rule-based evaluation benchmark specifically designed to systematically assess fine-grained compositional capabilities, such as object counting, attribute binding, and precise spatial relations.

\textbf{MS-COCO 2017}~\cite{lin2015microsoftcococommonobjects}: A standard benchmark extensively used for zero-shot text-to-image evaluation. We utilize its validation set prompts to measure overall image generation quality and general text-image alignment under standard descriptive conditions.

\textbf{ChronoMagic-Bench-150}~\cite{yuan2024chronomagic}: A comprehensive benchmark designed to assess temporal consistency, motion naturalness, and prompt adherence in text-to-video synthesis.

\subsection{Evaluation Metrics}

\textbf{HPSv2}~\cite{wu2023human} and \textbf{ImageReward}~\cite{xu2023imagerewardlearningevaluatinghuman}: Automated metrics trained on large-scale human feedback to measure how well the generated images align with human aesthetic preferences and intent.

\textbf{Aesthetic Score}~\cite{schuhmann2022laion5bopenlargescaledataset}: Used to evaluate the overall visual appeal and artistic quality of the outputs using a pre-trained aesthetic predictor.

\textbf{FID}~\cite{Seitzer2020FID} (Fréchet Inception Distance): Computes the distance between feature vectors calculated for real and generated images, quantifying foundational image fidelity and diversity relative to the real data distribution.

\textbf{CLIP Score}~\cite{radford2021learningtransferablevisualmodels}: Strictly quantifies the semantic consistency between the generated images (or video frames) and the conditioning text prompts.

\textbf{Top-$\boldsymbol{k}$ Accuracy}~\cite{NIPS2012_c399862d}: Evaluates the robust success rate of the generated images being ranked as the most preferred and highly aligned outputs within candidate sets.

\textbf{Video-Specific Metrics}: For temporal tasks, we evaluate motion consistency using \textbf{CHScore Flow}~\cite{yuan2024chronomagic}, frame-level structural preservation via \textbf{Frame LPIPS}~\cite{zhang2018perceptual} and \textbf{Frame SSIM}~\cite{nilsson2020understandingssim}, and spatial-temporal semantic alignment using \textbf{CLIP SIM}~\cite{radford2021learningtransferablevisualmodels} and \textbf{MTScore CLIP}.

\bibliographystyle{plainnat}
\bibliography{software}
\end{document}